\documentclass[runningheads]{llncs}
\usepackage{graphicx}

\usepackage{tikz}
\usepackage{comment}
\usepackage{amsmath,amssymb} %
\usepackage{color}

\usepackage[accsupp]{axessibility}  %

\usepackage{graphicx}
\usepackage{amsmath}
\usepackage{booktabs}
\usepackage{makecell}
\usepackage{subfig}
\usepackage{cuted}
\usepackage{tabularx}
\usepackage{multirow}
\usepackage[ruled,linesnumbered]{algorithm2e}

\usepackage[colorlinks]{hyperref}       %

\usepackage[capitalize]{cleveref}
\crefname{section}{Sec.}{Secs.}
\Crefname{section}{Section}{Sections}
\Crefname{table}{Table}{Tables}
\crefname{table}{Tab.}{Tabs.}

\newcommand{\fig}{Figure}

\newcommand{\as}[1]{}
\newcommand{\PreserveBackslash}[1]{\let\temp=\\#1\let\\=\temp}
\newcolumntype{C}[1]{>{\PreserveBackslash\centering}p{#1}}
\newcolumntype{R}[1]{>{\PreserveBackslash\raggedleft}p{#1}}
\newcolumntype{L}[1]{>{\PreserveBackslash\raggedright}p{#1}}

\def\eg{\emph{e.g.}}

\def\ie{\emph{i.e.}}

\def\etal{\emph{et al.}\ }

\DeclareMathOperator{\Gm}{\mathcal{G}}
\DeclareMathOperator{\W}{\mathbf{W}_{\Gm'}}
\DeclareMathOperator{\F}{F_{\Gm}}
\DeclareMathOperator{\Fenc}{F_{\text{enc}}}
\DeclareMathOperator{\FPool}{F_\text{pool}}
\DeclareMathOperator{\FLine}{F_\text{Line}}
\DeclareMathOperator{\E}{E_{\Gm}}
\DeclareMathOperator{\LE}{\mathcal{L_\text{E}}}
\DeclareMathOperator{\Wavg}{\overline{W}_{\Gm'}}
\DeclareMathOperator{\Lg}{\Gm_\text{Line}}
\DeclareMathOperator{\Eenc}{E_{\text{enc}}}
\DeclareMathOperator{\EPool}{E_\text{pool}}
\DeclareMathOperator{\ELine}{E_\text{Line}}

\begin{document}
\pagestyle{headings}
\mainmatter
\def\ECCVSubNumber{6187}  %

\title{Neural Space-filling Curves} %

\titlerunning{Neural Space-filling Curves}
\author{Hanyu Wang\inst{} \and
Kamal Gupta\inst{} \and
Larry Davis\inst{} \and
Abhinav Shrivastava\inst{}
}

\authorrunning{H. Wang et al.}
\institute{University of Maryland, College Park\newline
\email{\{hywang66@, kampta@cs., lsd@umiacs., abhinav@cs.\}umd.edu}\\
}

\maketitle

\begin{abstract}
   We present Neural \textbf{S}pace-\textbf{f}illing \textbf{C}urves (SFCs), a data-driven approach to infer a context-based scan order for a set of images. Linear ordering of pixels forms the basis for many applications such as video scrambling, compression, and auto-regressive models that are used in generative modeling for images. Existing algorithms resort to a fixed scanning algorithm such as Raster scan or Hilbert scan. Instead, our work learns a spatially coherent linear ordering of pixels from the dataset of images using a graph-based neural network. The resulting Neural SFC is optimized for an objective suitable for the downstream task when the image is traversed along with the scan line order. We show the advantage of using Neural SFCs in downstream applications such as image compression. Project page: \url{https://hywang66.github.io/publication/neuralsfc}.
\end{abstract}

\section{Introduction}

In any form of digital communication, information is transmitted via a sequence of discrete symbols. This includes images and videos, even though they are inherently signals with two spatial dimensions (2D). The modus operandi for transmitting such signals is to (1) efficiently encode and quantize their values in the spatial or spectral domain, (2) linearize the signal to a one-dimensional (1D) sequence by using a standard scanning order such as raster, zig-zag, or Hilbert Curve order~\cite{hilbert1935stetige}, and finally (3) apply a Shannon~\cite{shannon1948mathematical} style entropy coding technique such as Arithmetic coding~\cite{witten1987arithmetic} or Huffman coding~\cite{huffman1952method} to further compress the 1D sequence. Given the ubiquity of images and videos in our lives, a large amount of effort has gone into optimizing each of these steps of digital communication.  The focus of this work is the second step of linearizing the 2D spatial signal to a 1D sequence. A continuous scan order that traverses all spatial locations in two or higher dimensional signals exactly once is also known as the space-filling curve (SFC)~\cite{peano1890courbe}.

\begin{figure}[!h]
\centering
\includegraphics[width=0.9\linewidth]{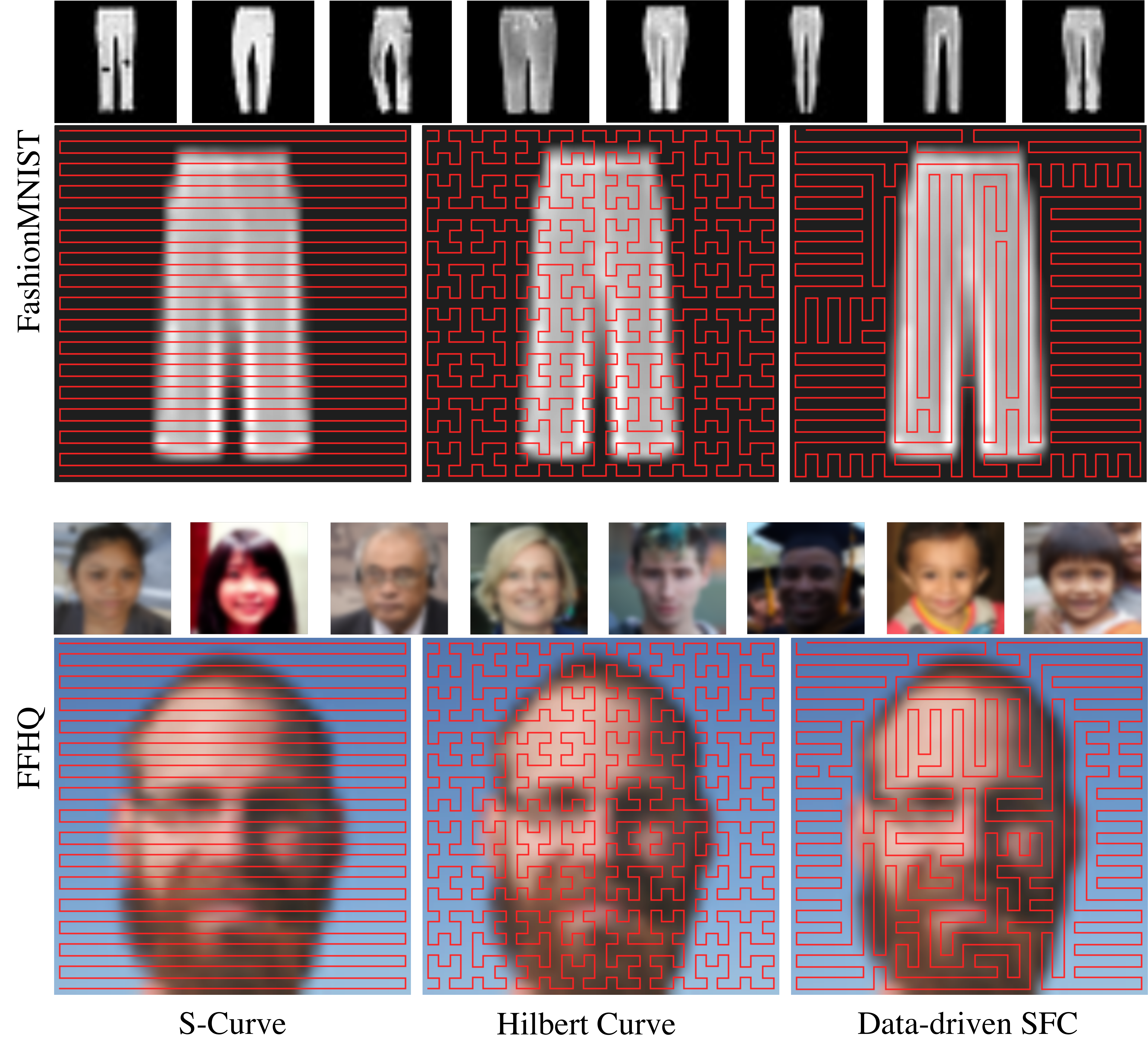}
\caption{Given a set of images, a gif, or a video, Neural Space-filling Curves (SFC) can provide a more spatially coherent scan order for images as compared to universal scan orders such as S-curve, or Peano-Hilbert curves. As shown in the example of a trouser and a face, the scan line tends to cover the background before moving to the foreground. (SFCs generated here using half-resolution images and resized for clarity. Best viewed in color.)}
\label{fig:teaser}
\end{figure}

Prior works have proposed various space-filling curves (SFCs) in the last hundred years, most of them context-agnostic, \ie, they are completely defined by the size and dimension of the space without taking into account spatial information of the space, \eg, pixels in the case of two-dimensional images. These universal context-agnostic SFCs are typically defined recursively to ensure simplicity and scale. Some of the SFCs also have spatial coherence properties and have been used in various image-based applications~\cite{moon2001analysis,kamata1993implementation,ansari1992image,thyagarajan1991fractal}. 

However, in many applications such as video conferencing, health-care, or social media, the images being transmitted, are often repetitive with a similar layout and content with minor variations transmitted over and over again. GIFs are another great example that consists of highly repetitive content and need to be stored efficiently and often, losslessly. Since universal SFCs do not utilize the intrinsic information of image content, they are far from optimal for a single image or a set of images with repetitive structure (refer to \cref{fig:teaser} for an example). Dafner \etal\cite{dafner2000context} proposed SFCs that exploit the inherent correlation between pixel values in an image.  Our work improves upon Dafner \etal, in two aspects.

\begin{itemize}
    \item Instead of discovering a single SFC for every image independently, we propose a data-driven technique to find optimal SFCs for a set of images. We postulate that context-based SFCs are more suitable for linearizing a group of images (or a short video/gif), since the cost of storing the SFC itself can be amortized by the number of images.
    \item We devise a novel alternating minimization technique to train an SFC weights generator, which allows us to optimize for any given objective function, even when not differentiable.
\end{itemize}

To the best of our knowledge, ours is the first work to propose a machine learning method for computing context-based SFCs and opens new directions for future research on optimal scanning of 2D and 3D grid-based data structures such as images, videos, and voxels. We demonstrate both quantitatively and qualitatively the benefit of our approach in various applications.

\section{Related Work}

Space-filling Curves (SFCs), introduced by Peano in 1890~\cite{peano1890courbe} and Hilbert in 1891~\cite{hilbert1935stetige}, are injective functions that map a line segment to a continuous curve in the unit square, cube, or hypercube. Most classic SFCs such as Peano-Hilbert, Sierpinski~\cite{sierpinski1912nouvelle},  and Moore~\cite{moore1900certain} curves are defined recursively which allows them to scale up to arbitrary resolution with a number of favorable properties.
In fact,~\cite{lempel1986compression} showed that the entropy of this pixel sequence asymptotically converges to the two-dimensional entropy of the original image for a large number of images coming from sufficiently random sources. Because of their self-organizing capabilities, Hilbert curves have found applications in compression~\cite{moon2001analysis,kamata1993implementation,ansari1992image,thyagarajan1991fractal}, computing~\cite{bader2012space}, recognition~\cite{alexandrov1982recursive,lee1994texture}, security~\cite{matias1987video}, databases~\cite{lawder2000calculation}, electronics~\cite{zhu2003bandwidth}, biology~\cite{lieberman2009comprehensive} and even web-comics~\cite{xkcd}. Hilbert curves have also been used to solve multidimensional task allocation problems in parallel processing~\cite{drozdowski2009scheduling}. This scheme is used in many job schedulers, such as the famous SLURM~\cite{yoo2003slurm}.

Dafner~\etal\cite{dafner2000context} proposed the first context-based SFCs. Their work makes use of cover and merge algorithm~\cite{matias1987video} to compute SFC for an image. Ouni \etal\cite{ouni2011gradient} employ image gradient based method to compute context-based SFCs, and they apply their SFCs to lossless compression tasks.
Zhou~\etal\cite{zhou2020data} improve Dafner~\etal's method by considering both data values and location coherency. They also generalize their method to multiscale data via quadtrees and octrees.

However, computing a unique SFC specific to an image has limited applications. Compression techniques such as LZW~\cite{lempel1986compression} that exploit the correlation between nearby pixel values during encoding, for example, can do a better job with context-based SFCs, however, the cost of storing an SFC itself for each image gives away the advantage. Universal SFCs, such as Hilbert curves don't have this disadvantage and are frequently used in compression applications. In this work, we argue that, applications that need to store and transmit images with repetitive structures such as gifs, can benefit greatly from context-based SFC optimized for that set of images.

Finding an optimal SFC is a combinatorial optimization problem. Solving combinatorial problems have a rich history in the field of machine learning. In 1985, ~\cite{hopfield1985neural} first attempted to solve these problems using a neural network. In the deep learning era, various flavors of attention have been proposed~\cite{vinyals2015pointer,vinyals2015order,deudon2018learning} to reach an approximate solution of NP-hard problems in computer science. Since combinatorial optimization problems are inherently non-differentiable, reinforcement learning (RL) is also a promising alternative for addressing these problems~\cite{vinyals2015order,dai2017learning,deudon2018learning,kool2018attention}. In our initial experiments, we found the performance of RL approach in our use-case to be unstable and inconsistent. %

\begin{figure*}[!t]

\begin{minipage}{.2\linewidth}
\centering
\subfloat[\footnotesize{Cover $\Gm$}]{\label{fig:cover}\includegraphics[width=\textwidth]{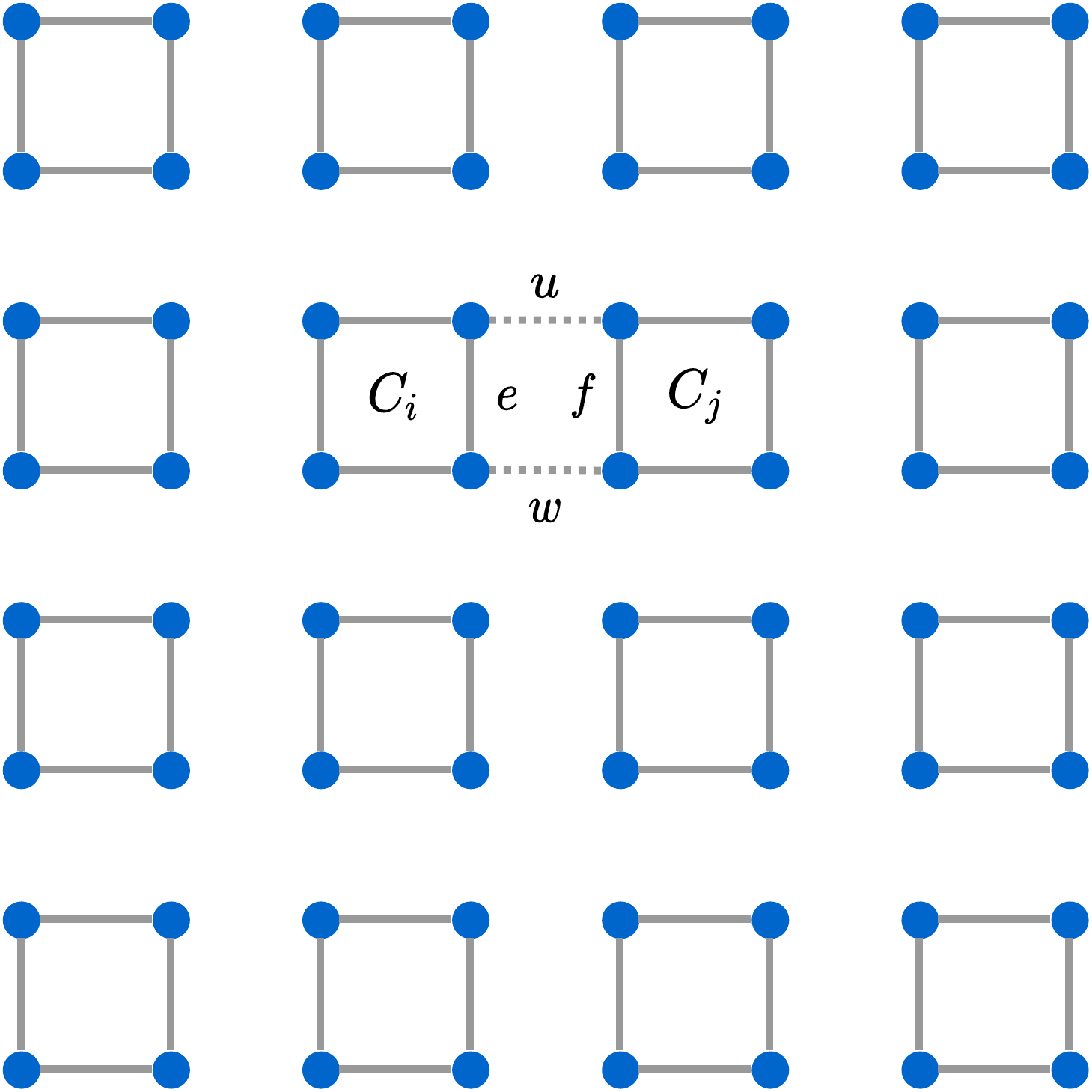}}
\end{minipage} \hfill
\begin{minipage}{.2\linewidth}
\centering
\subfloat[\footnotesize{Dual $\mathcal{G'}$}]{\label{fig:dual}\includegraphics[width=\textwidth]{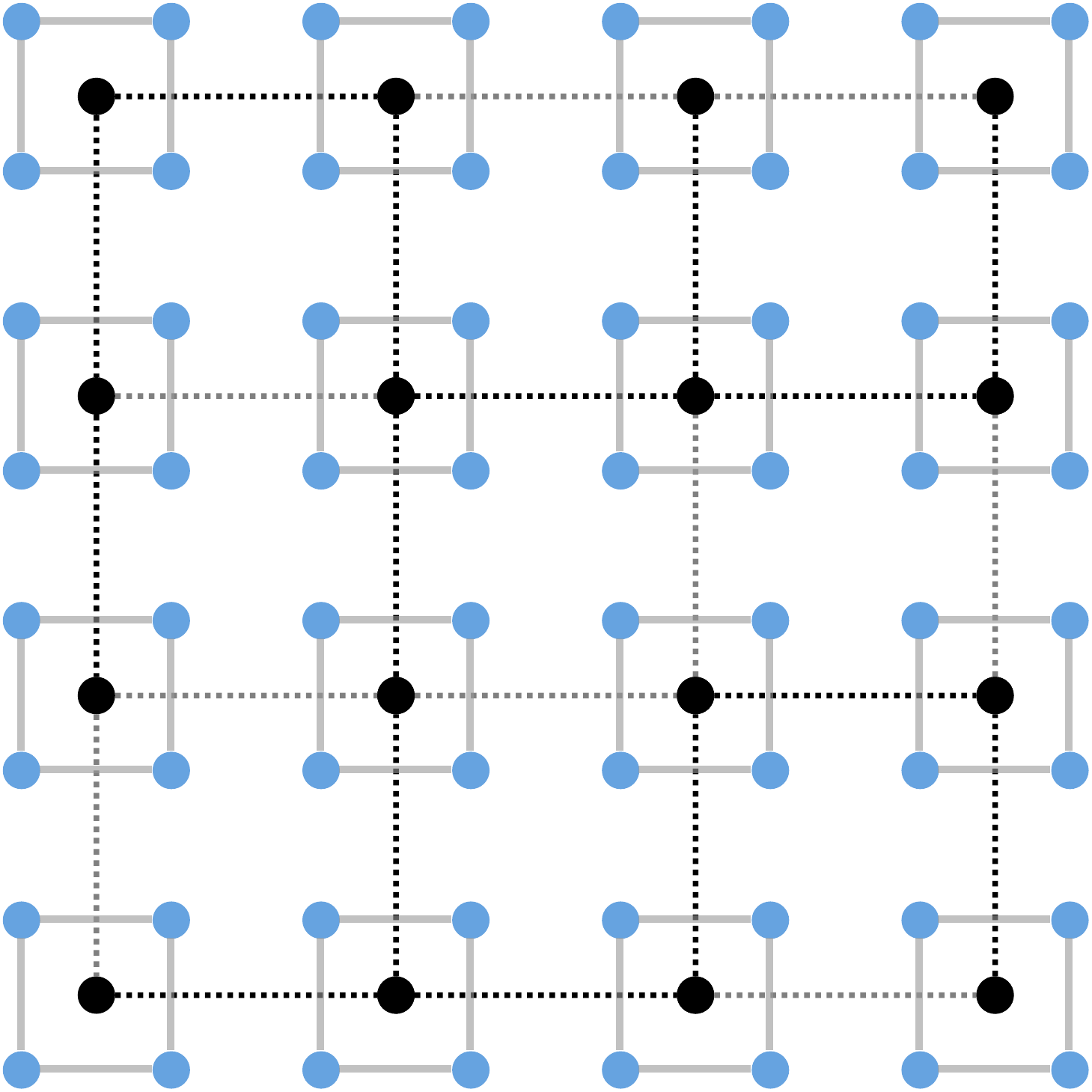}}
\end{minipage} \hfill
\begin{minipage}{.2\linewidth}
\centering
\subfloat[\footnotesize{$\mathcal{T}$: MST of $\mathcal{G'}$}]{\label{fig:mst}\includegraphics[width=\textwidth]{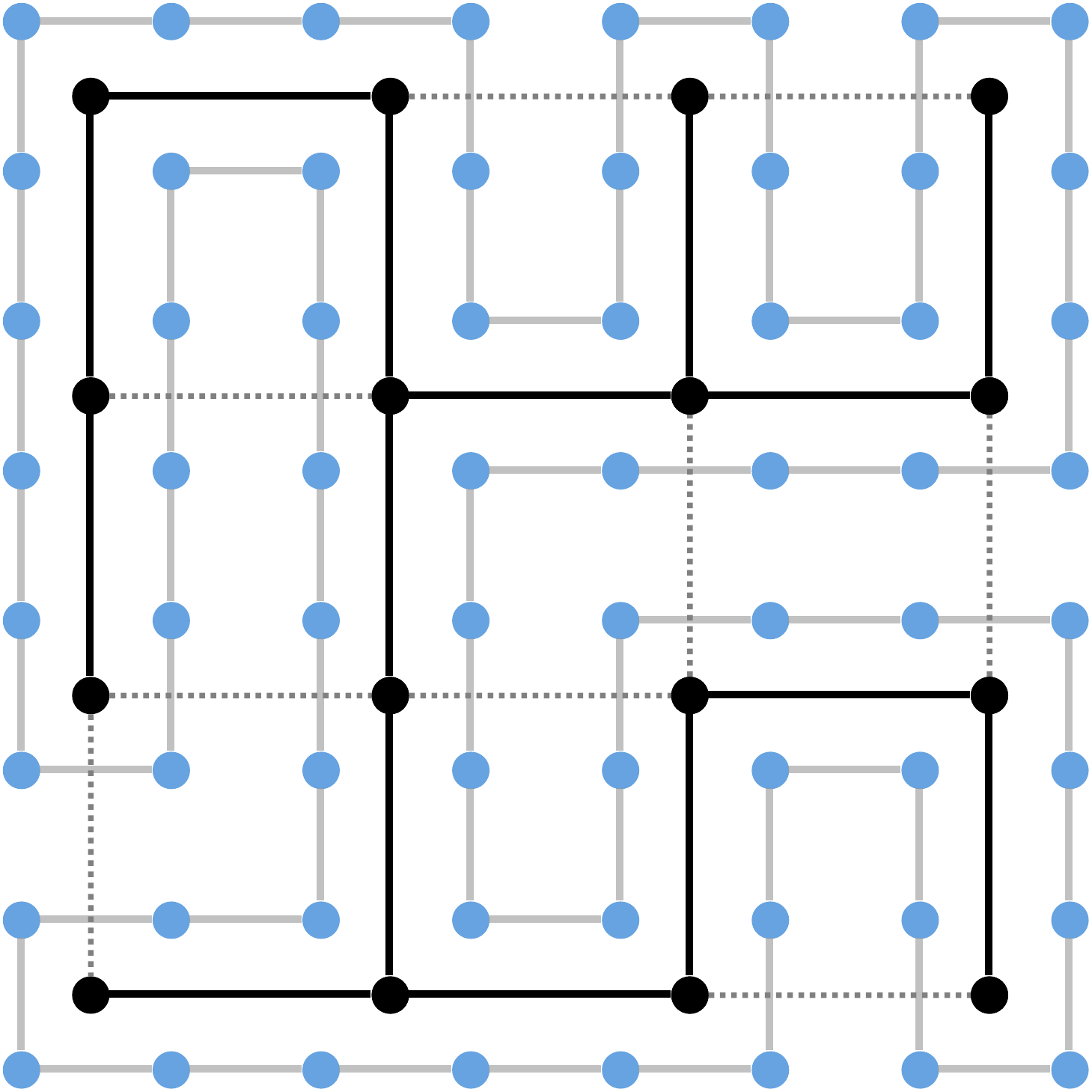}}
\end{minipage} \hfill
\begin{minipage}{.2\linewidth}
\centering
\subfloat[\footnotesize{Merged}]{\label{fig:merged}\includegraphics[width=\textwidth]{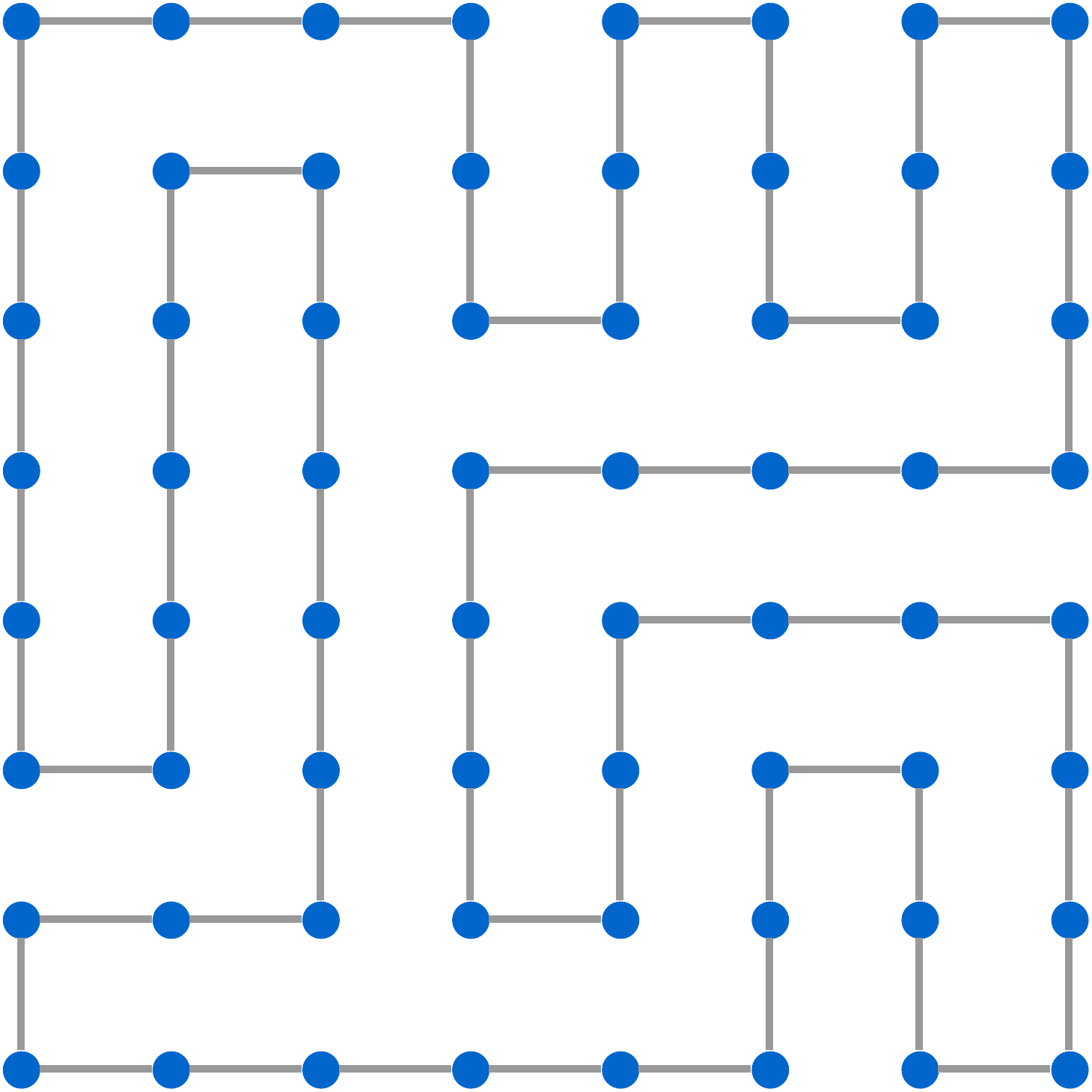}}
\end{minipage}
\caption{\textbf{Cover and Merge Algorithm.} (a): An $8 \times 8$ image fully covered by $2 \times 2$ circuits. (b): The dual graph $\Gm'$ (black) built on the covering circuits $\Gm$ (blue). (c): The Minimum Spanning Tree $\mathcal{T}$ (black solid lines) of $\Gm'$ (all black lines) and the Hamiltonian Circuit (blue) induced by $\mathcal{T}$. (d): A single Hamiltonian circuit merged from the covering circuits. See more details in \cref{sec:cbsfc}.}
\label{fig:main}
\end{figure*}

\section{Approach}
\label{sec:approach}

We first describe the algorithm for computing SFC for one image as proposed by Dafner~\etal\cite{dafner2000context} in \Cref{sec:cbsfc}. We then extend this treatment to a more general setting where we can optimize the SFC for any non-differential objective function for multiple images in \Cref{sec:nsfc}. The rest of the \Cref{sec:approach} describes major components of our framework and training procedure in detail.

\subsection{Overview of Dafner \etal (Single Image-based SFC)}
\label{sec:cbsfc}

Given an image,~\cite{dafner2000context} represents it as an undirected graph $\Gm$ whose nodes are pixel locations, and each pixel is connected to its 8 neighboring pixels by an edge. Generating a context-based SFC from the given image is then equivalent to finding a Hamiltonian path in graph $\Gm$.
They use the cover and merge algorithm (initially proposed by Matias \etal\cite{matias1987video} to scramble a video signal for secure transmission). As the name suggests, the algorithm works by finding a Hamiltonian path for the image-grid graph $\Gm$ in two steps - \textbf{cover} and \textbf{merge}.

In the \textbf{cover} step, a dual undirected graph $\Gm'$ is constructed from $\Gm$. The vertices of $\Gm'$ are small disjoint square circuits covering the whole of $\Gm$ as shown in \Cref{fig:cover}. Each circuit covers $4$ pixels. We call these initial circuits $C_1, C_2, ..., C_k$ and connect $(C_i, C_j)$ if the circuits $C_i$ and $C_j$ are adjacent in the original graph $\Gm$.
\Cref{fig:dual} shows a dual graph example built for an $8 \times 8$ image.

In the \textbf{merge} step, all circuits are merged to form a single Hamiltonian circuit. To merge the circuits optimally, a weight is assigned to each edge $(C_i, C_j)$ in $\Gm'$ representing the ``cost'' of merging circuits $C_i$ and $C_j$. 
The weight $w(C_i, C_j)$ of the edge connecting circuits $C_i$ and $C_j$ is defined as the cost of exchanging the edges $e$ and $f$ with the edges $u$ and $w$ in the image graph:
\begin{equation}
\label{eqn:dafner_weights}
w(C_i, C_j) = |u| + |w| - |e| - |f|,
\end{equation}
where $|\cdot|$ corresponds to the absolute difference in pixel values at the two vertices of the edge in $\Gm$. A minimum spanning tree $\mathcal{T}$ is then constructed using these weights. \Cref{fig:mst} shows what $\mathcal{T}$ might look like for $\Gm'$. Next, we start merging circuits that are part of $\mathcal{T}$. Merging two circuits corresponds to removing their adjacent edges in $\Gm$ (\eg, edge $e$ and $f$ in \Cref{fig:cover}), and creating new edges (\eg, edge $u$ and $w$ in \Cref{fig:cover}). Note that only adjacent circuit pairs can be merged. Given the spanning tree, all merging operations on $\Gm$ can be done in a linear time to obtain a Hamiltonian circuit as in \Cref{fig:merged}. Finally, the SFC or Hamiltonian path can be obtained by cutting the circuit at an arbitrary point.

\subsection{Neural Space-filling Curves}
\label{sec:nsfc}

\begin{figure*}[!t]
\centering
    \includegraphics[width=\linewidth, trim=45 0 40 0, clip]{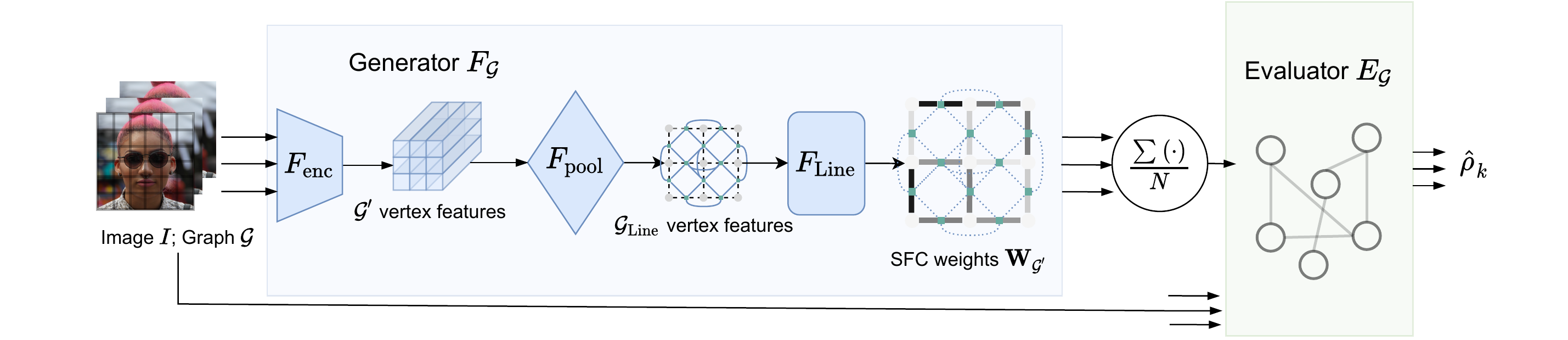}
    \caption{\textbf{Neural SFC pipeline.} The Neural SFC model consists of a weight generator $\F$ and a weight evaluator $\E$. Taking one or more images as input, $\F$ extracts the deep features of the image and generates the SFC weights. $\E$ takes both the image(s) and the corresponding SFC weights as input $\E$ then estimates the negative autocorrelation of the image pixel sequence inferred by the input weights to evaluate the goodness of the input weights with respect to the input image. See more details in \cref{sec:nsfc}.}
\label{fig:approach}
\end{figure*}

Dafner's approach~\cite{dafner2000context} is effective at exploiting the local relationships in an image. However, it has a few limitations. 
\begin{itemize}
    \item Since the edge weights are computed using only two adjacent pixels, the receptive field is limited, and the resulting SFC does not take into account the long-range context in the image.
    \item Dafner's approach only works for one image at a time. The notion of context-based SFCs can be further generalized to finding an SFC for a set of images.
    \item In the current form, \cite{dafner2000context} cannot optimize for arbitrary objective functions. The context-based SFC obtained is closely tied to edge weights defined by Eq.~\ref{eqn:dafner_weights} which encourage autocorrelation with lag-2 in the pixel sequence obtained.
\end{itemize}

We address each of these issues in our data-driven approach to infer an optimal SFC for a set of images for any objective function. For brevity, we will optimize for lag-$k$ autocorrelation, however, in our experiments, we will show how we can modify this objective and directly optimize for a downstream application such as compression.

\medskip
\noindent
\textbf{Setup.} For the remainder of this section, we use the following notation. We are given a set of $N$ images. Each image (color or grayscale) has a resolution of $H \times W$. For a given image $I$, we define graph $\Gm$ over its $HW$ pixel locations. The dual graph $\Gm'$ consists of $2\times2$ disjoint circuits covering all the vertices of $\Gm$ as defined in the previous section. We first note that the whole cover and merge algorithm is context-agnostic barring the step where we assign weights to the edges of the dual graph. Therefore, a context-based SFC can be completely parameterized by these weights, denoted by $\W$. For a given set of weights of image $I$, the merge operation provides us with a Hamiltonian circuit. A Hamiltonian path, or an SFC, can be obtained by breaking the merged Hamiltonian circuit at any point.  Hence, the problem of finding an SFC can be reduced to finding the optimal weights $\W$. 
We propose to learn a neural network, $\F$, to approximate $\W$.
Once we have the optimized weights, the merge operation is fast and efficient in terms of both memory and speed, and we exploit it to get our desired SFC.

\subsection{Weight Generator}
\label{sec:wg}
The weight generator $\F$ is designed to take as input a single image $I$,
and output the edge weights $\W$ of its dual graph, \ie, $\W = \F(I)$.
While the input dimension is $H \times W$, the output $\W$, has a size equal to the number of edges in the dual graph $\Gm'$. It is trivial to show that for the dual graph $\Gm'$ consisting of $\frac{H}{2} \times \frac{W}{2}$ vertices (or circuits), the corresponding number of edges will be $\frac{HW-H-W}{2}$. Further, the edges of $\Gm'$ do not conform to a 2D grid structure as the input image $I$. 
To address this, we decompose the weight generator $\F$ further in three modules.
\begin{align}
\W = \F(I) = \FLine \circ \FPool \circ \Fenc (I).
\label{eq:generator}
\end{align}
\Cref{fig:approach} shows the architecture of the weight generator. The first submodule is a dual graph encoder $\Fenc$, which takes $I$ as the input and extracts a deep representation of the vertices of dual graph $\Gm'$, resulting in a $\frac{H}{2} \times \frac{W}{2} \times d$ dimensional output, where $d$ is the number of output feature maps. In this work, $\Fenc$ is implemented using a fully convolutional neural network with residual connections.

The second submodule $\FPool$ consists of two pooling filters (and no trainable weights) of dimension $1\times 2$ and $2 \times 1$ applied sequentially. $\FPool$  imitates the graph operations by aggregating the features of vertices, and hence forming edge features of $\Gm'$.
Given a $d$-dimensional representation of edges, we want to compute a scalar weight for each edge in $\Gm'$. It is desirable that this scalar weight can exploit not only the edge features, but also long-range relationships among the edges. Hence we construct a Line graph~\cite{harary1960some} $\Lg$ to represent the adjacency between edges of $\Gm'$. Each vertex in $\Lg$ corresponds to an edge in $\Gm'$. For every two edges in $\Gm'$ that have a vertex in common, there is an edge between their corresponding vertices in $\Lg$.

Using the Line graph, the edge features of $\Gm'$ becomes the vertex feature of $\Lg$, and the adjacent relations of edges of $\Gm'$ are represented by edges of $\Lg$. To compute the scalar weights on $\Lg$, we introduce the third submodule, a weights regressor $\FLine$ to run on $\Lg$.
$\FLine$ can be implemented using Graph Neural Network modules. In this work, we use GCN~\cite{kipf2016semi} in MNIST experiments and GAT~\cite{vel2018graph} in FFHQ Faces experiments.

\subsection{Objective Functions}
\label{ssec:objective}
The weight generator described above can generate edge weights for a given image. For every mini-batch of images, we take an expected value of the weights for all the images to get $\Wavg$. Given $\Wavg$, we can compute an SFC for the mini-batch of images. The quality or `goodness' of an SFC can be different for different applications. 
In this paper, we consider two plausible objectives: the 1D autocorrelation and the LZW sequence length. For both of them, the first step is to flatten the given image $I$ to the 1D pixel sequence $\{y_i\}$ based on the SFC defined by $\Wavg$.

\medskip
\noindent\textbf{Autocorrelation.}
The 1D autocorrelation measures the internal local similarity of a 1D sequence. Therefore, the smaller the 1D autocorrelation is, the better the SFC is.
The lag-$k$ 1D autocorrelation of a pixel sequence $\{y_i\}$ of length $HW$ is defined as
\begin{align}
    \rho_{k} = \frac{\sum_{i=1}^{HW-k}y_{i}y_{i+k} } {\sum_{i=1}^{HW}y_{i}^{2} }.
\label{eq:autocorrelation}
\end{align}

\medskip
\noindent\textbf{Code Length.}
Another SFC quality metric, the LZW sequence length, is inspired from the Lempel-Ziv Welch (LZW) encoding~\cite{ziv1978compression,welch1984technique}, which is popularly used to encode GIFs losslessly. Its performance depends on the amount of redundant data in a given sequence.
Given a pixel sequence $\{y_i\}$, it's LZW length is defined as 
\begin{align}
    L = \mathrm{length}(\mathrm{Encode}(\{y_i\})),
\label{eq:lzwl}
\end{align}
where 
$\mathrm{Encode}$ is the LZW-encoding function, and $\mathrm{length}$ measures the length of a sequence. 

Note that computing $\rho_{k}$ or $L$ requires us to first obtain a minimum spanning tree to infer the SFC from $I$ and $\Wavg$, which is a non-differentiable operation.
Therefore, these metrics, by themselves, cannot be directly used to optimize a neural network. Hence we can't simply backpropagate gradient information to update $\F$.

To overcome the problem of non-differentiability of SFC computation, we train an evaluator neural network, $\E$, as a differentiable proxy to estimate the resulting autocorrelation of SFC weights computed by $\F$ from the context image(s). By carefully designing the training procedure, our model $\E$ is able to approximate any non-differentiable metric, hence serving as an effective loss function of the weight generator. \Cref{fig:approach} summarizes our approach.

Also note that, while we refer to the lag-$k$ autocorrelation and the LZW length as our objective functions, we can replace them with any loss (or reward) suitable for a given application, even if it is not differentiable.

\subsection{Weight Evaluator}

The weight evaluator, denoted by $\E$, acts as a differentiable proxy to estimate the objective $\Phi$. Depending on the task, we choose 
$\Phi = -\rho_k$
for minimizing the negative autocorrelation or
$\Phi = L$
for minimizing LZW length.

Given the average SFC weights $\Wavg$ and an image $I$, we compute
\begin{align}
\hat{\Phi} &= \E(\Wavg, I), \\
\LE &= \mathbb{E} \left[\| \Phi -\hat{\Phi} \| \right], 
\label{eq:l2}
\end{align}
where $\LE$ denotes the expected value of $l$-2 error between groundtruth lag-$k$ autocorrelation and the predicted autocorrelation by $\E$ computed for the entire batch. $\LE$ serves as the objective function for training the evaluator.

The inputs to the weight evaluator $\E$ are $\Wavg$ and $I$. 
Similar to ~\cref{sec:wg}, we again decompose the weight evaluator $\E$ into submodules, 
\begin{align}
\hat{\Phi} = \E(\Wavg, I) = \ELine(\Wavg \| \EPool \circ \Eenc (I)),
\label{eq:rhok}
\end{align}
where $\|$ denotes the concatenation along the feature dimension. Following the graph encoding procedure of the weight generator $\F$, $\EPool$ and $\Eenc$ are functionally identical to $\FPool$ and $\Fenc$, respectively. The final submodule $\ELine$, which takes similar input to $\FLine$, has a different head to predict the estimated objective value $\hat{\Phi}$. 
The backbone of $\ELine$ is also implemented using a GCN or GAT, followed by an average pooling operation and a simple MLP to predict $\hat{\Phi}$ as a single value.

\begin{algorithm}[!t]
    \SetAlgoLined
    \KwData{A set of $N$ images each of resolution $H\times W$}
    \KwResult{SFC weights $\Wavg$ for the image set}
    
    Randomly initialize $\F$ and $\E$\;
    \Repeat{$\LE \rightarrow 0$}{
        \tcp{training $\F$}
        Sample a minibatch of $B$ images\;
        Forward pass for the weight generator $\F$,  $\W \leftarrow \F(I)$\;
        Expected weights for the mini-batch $\Wavg \leftarrow \mathbb{E}(\W)$\;
        Forward pass for the weight evaluator $\hat{\Phi} \leftarrow \E(\Wavg, I)$\;
        SGD update for $\F$ keeping $\E$ fixed, $\nabla_{\F} \mathbb{E} \left[\hat{\Phi} \right] $\;
        \BlankLine
        \BlankLine
        \tcp{training $\E$}
        For each example in $B$, with a probability $p_1$, get $\W$ using Eq.~\ref{eqn:dafner_weights}, with a probability $p_2$, sample $\W \sim \mathcal{N}\left(0,\,1\right)$ and with a probability $1-p_1-p_2$, keep $\W = \F(I)$\;
        Run a forward pass for the weight evaluator $\hat{\Phi} \leftarrow \E(\Wavg, I)$\;
        For the whole batch, compute the ground truth $\Phi$ using Eq.~\ref{eq:autocorrelation} or Eq.~\ref{eq:lzwl}\;
        SGD update for $\E$ with $\nabla_{\E} \LE $\;
    }
    \caption{Training Neural SFC}
    \label{alg:training}
\end{algorithm}

\subsection{Training}
\label{sec:training}
We adopt an alternating optimization procedure for training the core components of our architecture $\F$ and $\E$. Algorithm~\ref{alg:training} gives an overview of the training schema. The weight evaluator $\E$ solves the regression task of computing the estimated objective $\hat{\Phi}$ for a given set of SFC weights $\Wavg$. Given the context image $I$ and SFC weights $\Wavg$, we can get the groundtruth objective by first running Prim's algorithm~\cite{prim1957shortest} to get the desired SFC followed by Eq.~\ref{eq:autocorrelation} or Eq.~\ref{eq:lzwl}. 

Once we have the groundtruth $\Phi$, we optimize a standard L2 loss commonly used in regression methods to train $\E$, as we described in Eq.~\ref{eq:l2}. Since we eventually need to use $\E$ to train the upstream network $F$, we would like $\E$ to be trained on a diverse range of input SFC weights $\W$. 

The $\F$ can be trained trivially using a fixed $\E$. We empirically observed that training them alternately improves the training dynamics thus boosting the SFC quality.

\section{Experiments}
In this section, we evaluate the ability of the proposed training scheme to generate optimized Space-filling Curves (SFC) for a set of images. We further validate the efficacy of the Neural SFCs on real-world applications such as image or gif compression. We compare with standard Raster scan and Hilbert curves. Note that even though Raster scan is not an SFC mathematically speaking, we use it for benchmarking in our experiments due to its prevalence.

\subsection{Datasets} 

We trained the Neural SFC model on four different datasets. Both \textbf{MNIST}~\cite{lecun1998gradient} and \textbf{Fashion-MNIST}~\cite{xiao2017fashion} comprise of 60000 training images, and 10000 test images. Each image is a $28\times28$ grayscale image, which we resize to $32\times32$ to do a fair comparison with Hilbert curves which can be defined only when the image resolution is a power of 2. Resizing is done by a simple zero padding around the image. We observe a lot of intra-class similarity in the case of both MNIST and Fashion-MNIST, \ie, the images within the same class are similar in layout and content to each other, and hence we train a single SFC for each MNIST and Fashion-MNIST class. 

We also consider \textbf{FFHQ}~\cite{karras2019style} dataset. We downsample all the images to size $32 \times 32$ using bilinear interpolation. 
FFHQ is a dataset of celebrity faces and contains less noise compared to datasets like CelebA~\cite{liu2015faceattributes}. We split the dataset into 60000 training and 10000 test images. We train a single SFC for all the images in the data.

Lastly, in order to demonstrate a real-world application of SFCs designed for a set of images, we use a large-scale GIF dataset \textbf{TGIF}~\cite{li2016tgif}. The dataset consists of 80,000 training gifs, and 11,360 test gifs. 
We train a single Neural SFC model that takes a gif as an input and outputs an optimized SFC for the gif. We evaluate it on every gif in the test dataset. Average numbers are reported.

\subsection{Training details}

We consider two different objective functions for training Neural SFC. First, in order to compare our method with Dafner \etal\cite{dafner2000context}, we train our model using the autocorrelation objective function. Since Dafner's method cannot generate SFCs for an image-set trivially, we compute Dafner's image-set SFC by taking the expected value of $\W$ defined in their method for all the images in the training set. 
Another possible choice is to calculate an average image for the entire image set, then run Dafner's method on it to generate an image-set SFC. We use the first setting in all experiments in our main paper because it empirically performs better.
In all our experiments, we set $k=6$, such that lag-$6$ autocorrelation is used to train the weight evaluator $\E$.

Second, we also provide quantitative results on the compression of images and gifs using a Lempel-Ziv encoder (LZW) encoding scheme. Specifically, we optimize the SFC separately for each gif, although, it is possible to train a large SFC encoder on the entire gif dataset. We leave the study for evaluating context-based SFC for large- gifs or video datasets for future work.

\begin{figure*}[!ht]
\centering
    \includegraphics[width=\linewidth]{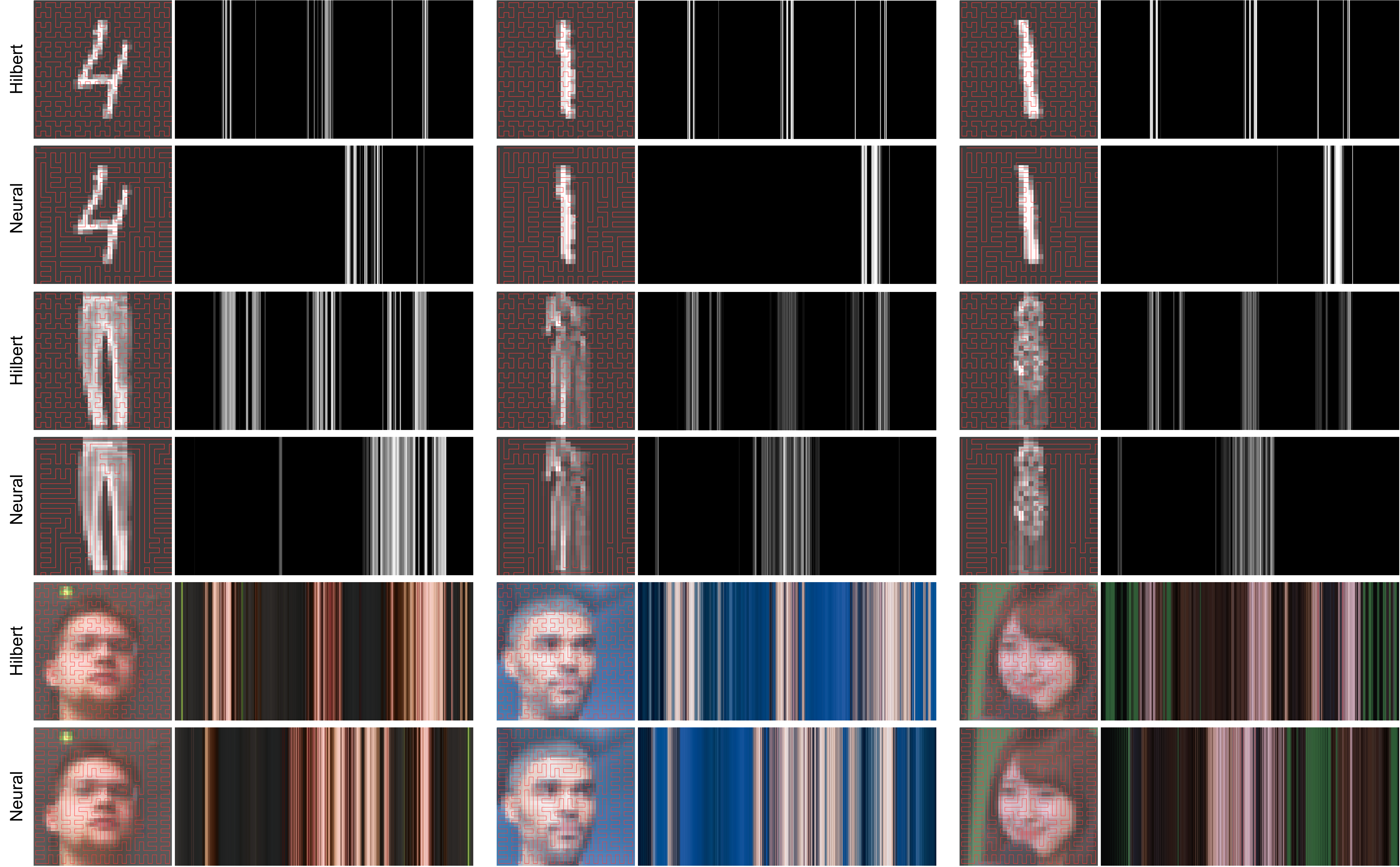}
    \caption{Qualitative comparison between Hilbert curves and Neural SFCs. Left: SFC (in red color) overlayed on the image. Right: Image flattened according to the SFC and visualized in 1-dimension. Images in the top two rows are from MNIST, the ones in the middle two rows are from Fashion-MNIST, and the ones in the bottom two rows are from FFHQ Faces. Neural SFCs on images from MNIST and Fashion-MNIST are class-conditional, \ie, computed for each class. Therefore, for MNIST and Fashion-MNIST, Neural SFCs in the right two columns are the same since the two images have the same class label.
    In all datasets, Neural SFCs are more spatially coherent and produce fewer clusters when visualized in 1-dimension. Best viewed in color.}
    \label{fig:vis_mnist_avg}
\end{figure*}

\subsection{Qualitative Evaluation}
\label{sec:qualeval}

\noindent\textbf{Visualizing the 1D sequence.}
In \Cref{fig:vis_mnist_avg}, we show a few examples to illustrate the difference between Neural SFCs and Hilbert Curve on the MNIST, Fashion-MNIST, and FFHQ images. 
We look at the SFC (red curve overlayed on the image of the digits/clothing/faces). While the Hilbert Curve outputs the same SFC for any $32 \times 32$ image in the world, Neural SFC optimizes the SFC for a given set of images.
To see this difference, we flatten the SFCs into a 1D sequence (images on the right of the digits/clothing/faces), and observe that Neural SFCs tend to keep better long-range spatial coherence than Hilbert Curve. In both cases, Neural SFCs show pixels in fewer clusters as compared to Hilbert curves.
Specifically, Neural SFCs are able to roughly stay in the bright regions until they all get covered. Therefore, bright pixels will mostly gather in one contiguous segment in the 1D sequence inferred from a Neural SFC. In contrast, Hilbert curves often result in multiple clusters of contiguous structures in the 1D sequence. 
We provide more examples in the supplementary material. 

\begin{figure}[!ht]
\centering %
\includegraphics[width=0.8\linewidth]{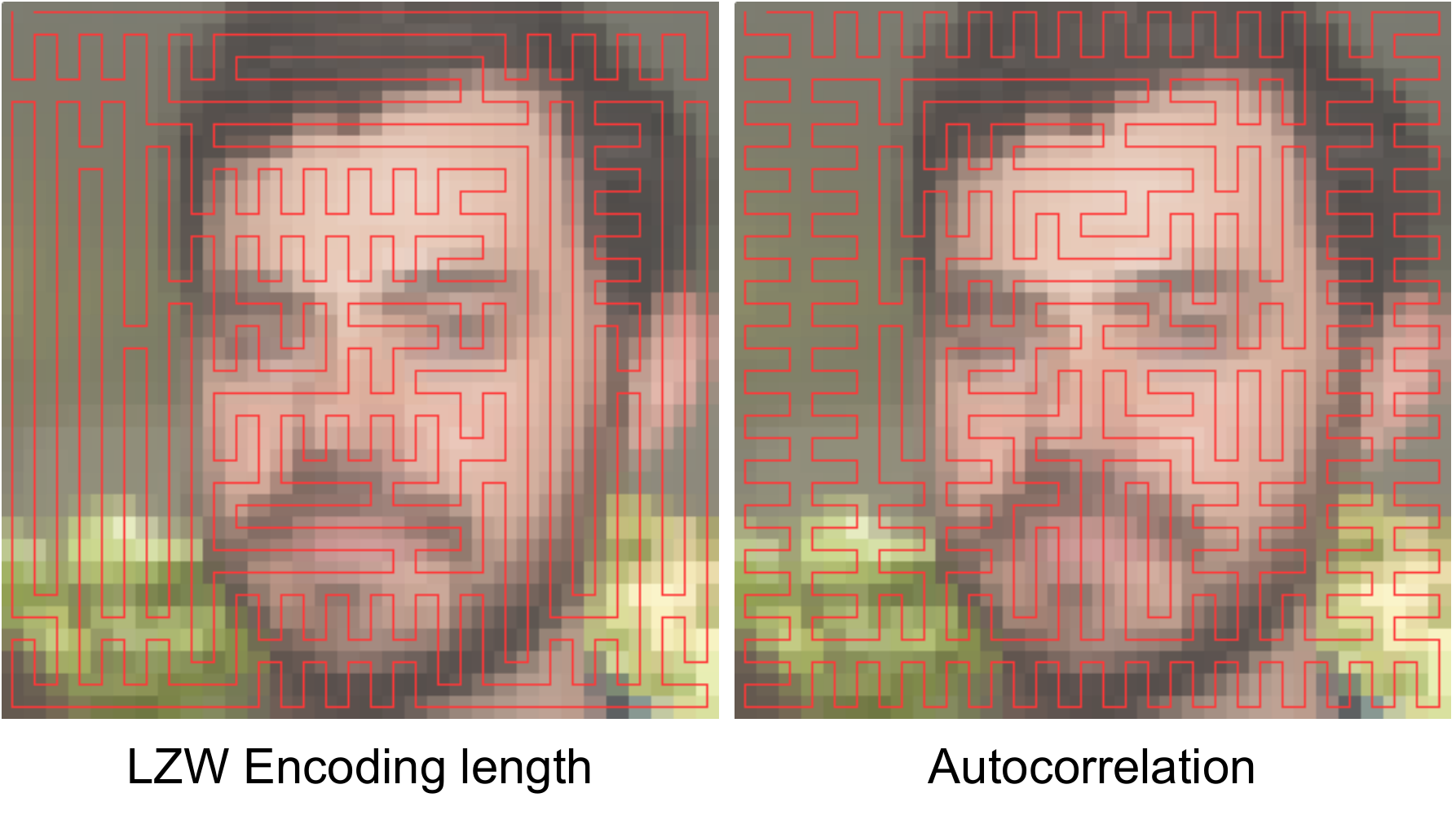}
\caption{We visualize the Neural SFCs training with two objectives considered in this paper -- autocorrelation and LZW encoding.}
\label{fig:ffhq_ac_lzw}
\end{figure}

\medskip
\noindent\textbf{SFCs obtained with different objectives.}
Fig.~\ref{fig:ffhq_ac_lzw} shows two different SFCs obtained by our approach when trained with different objective functions. The figure on the left corresponds to SFC optimized for LZW encoding length, while the figure on the right corresponds to SFC optimized for auto-correlation. We observe that generally, SFCs optimized for LZW encoding length are better at short-lag auto-correlation (as compared to SFCs optimized for lag-6 as in most of our experiments). This results in an SFC with fewer turns and straighter paths.

\begin{figure*}[!ht]
    \centering
\setlength{\tabcolsep}{1pt}
\begin{tabular}{ccc}
\includegraphics[width=0.33\linewidth]{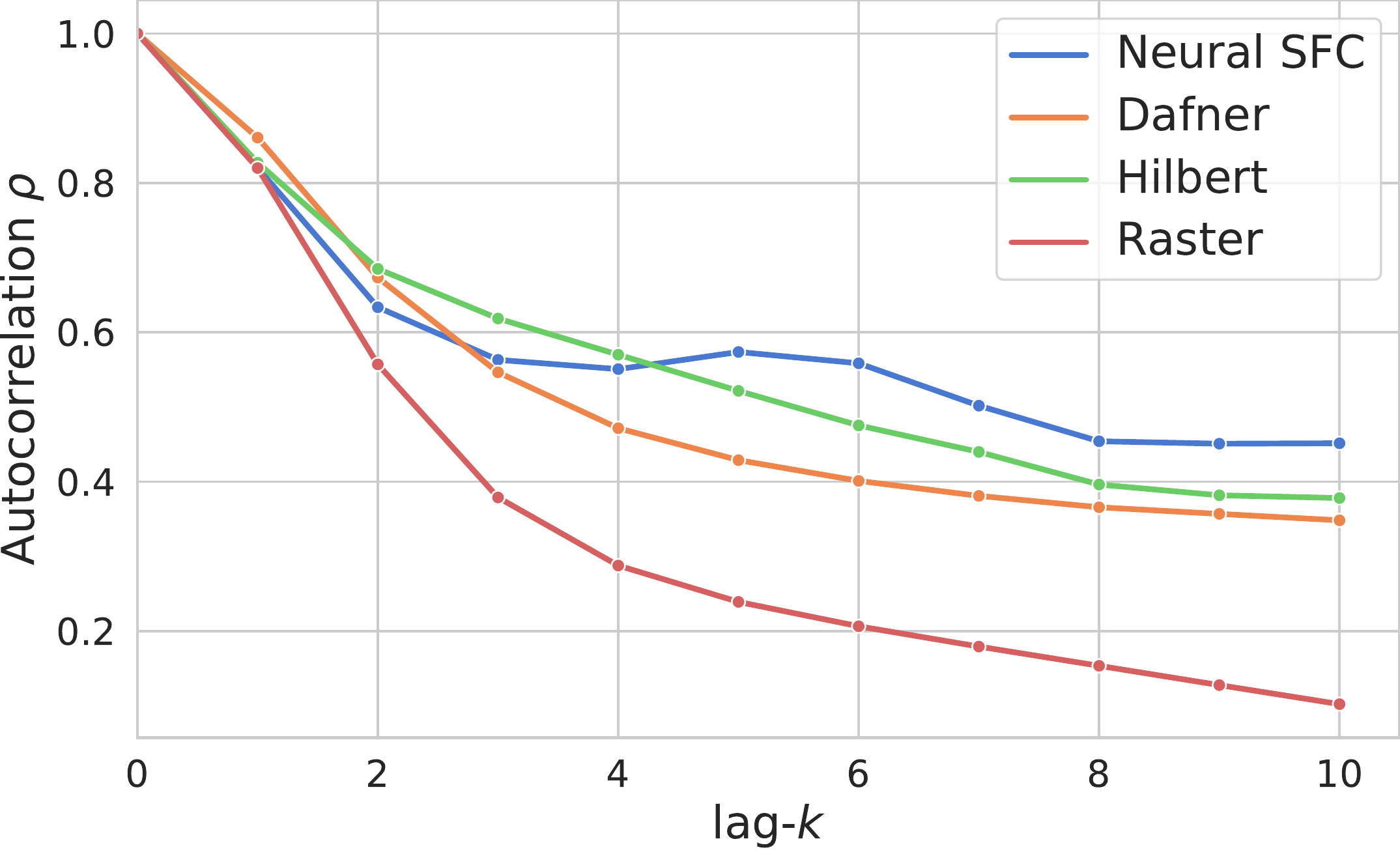}&
\includegraphics[width=0.33\linewidth]{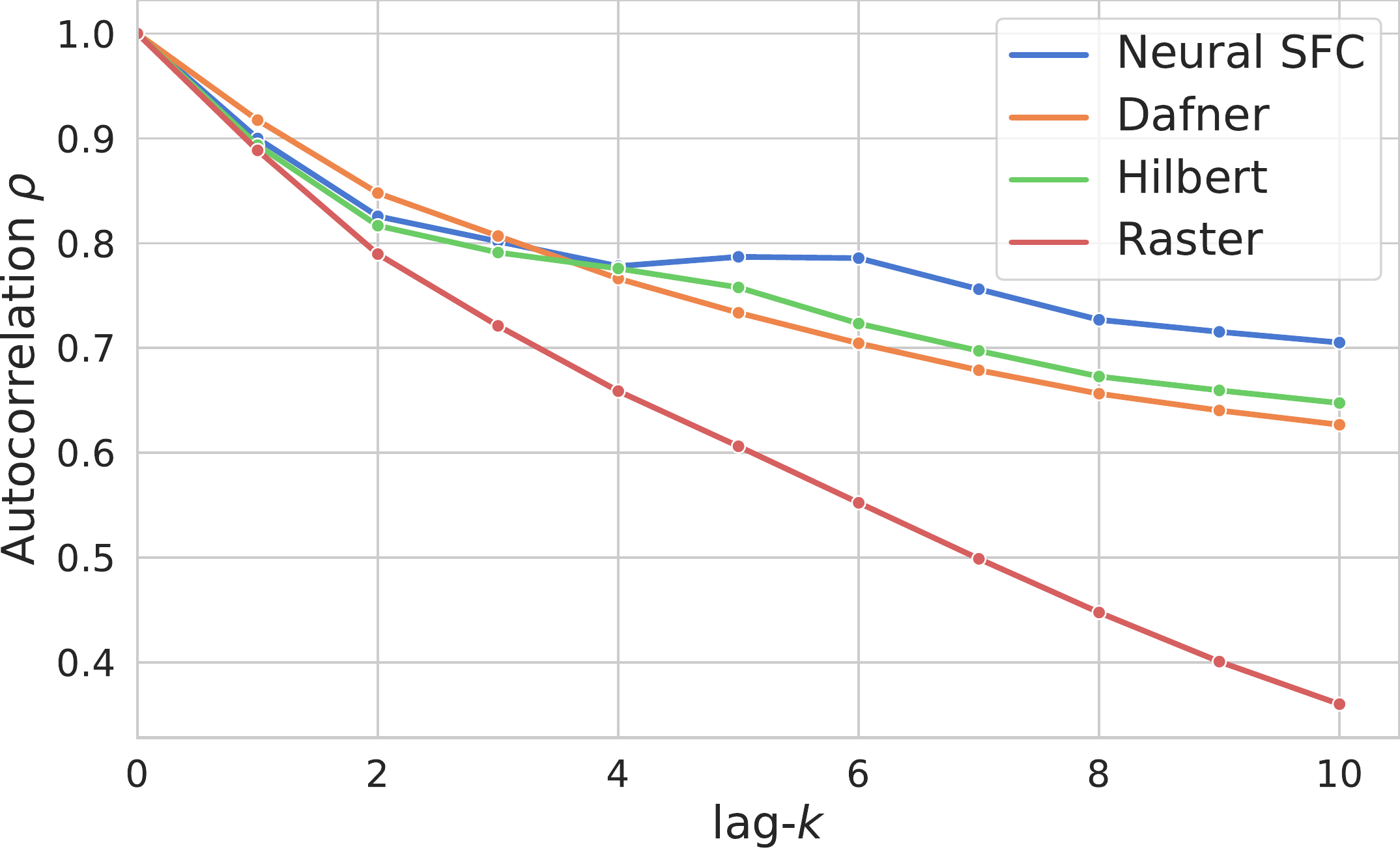}&
\includegraphics[width=0.33\linewidth]{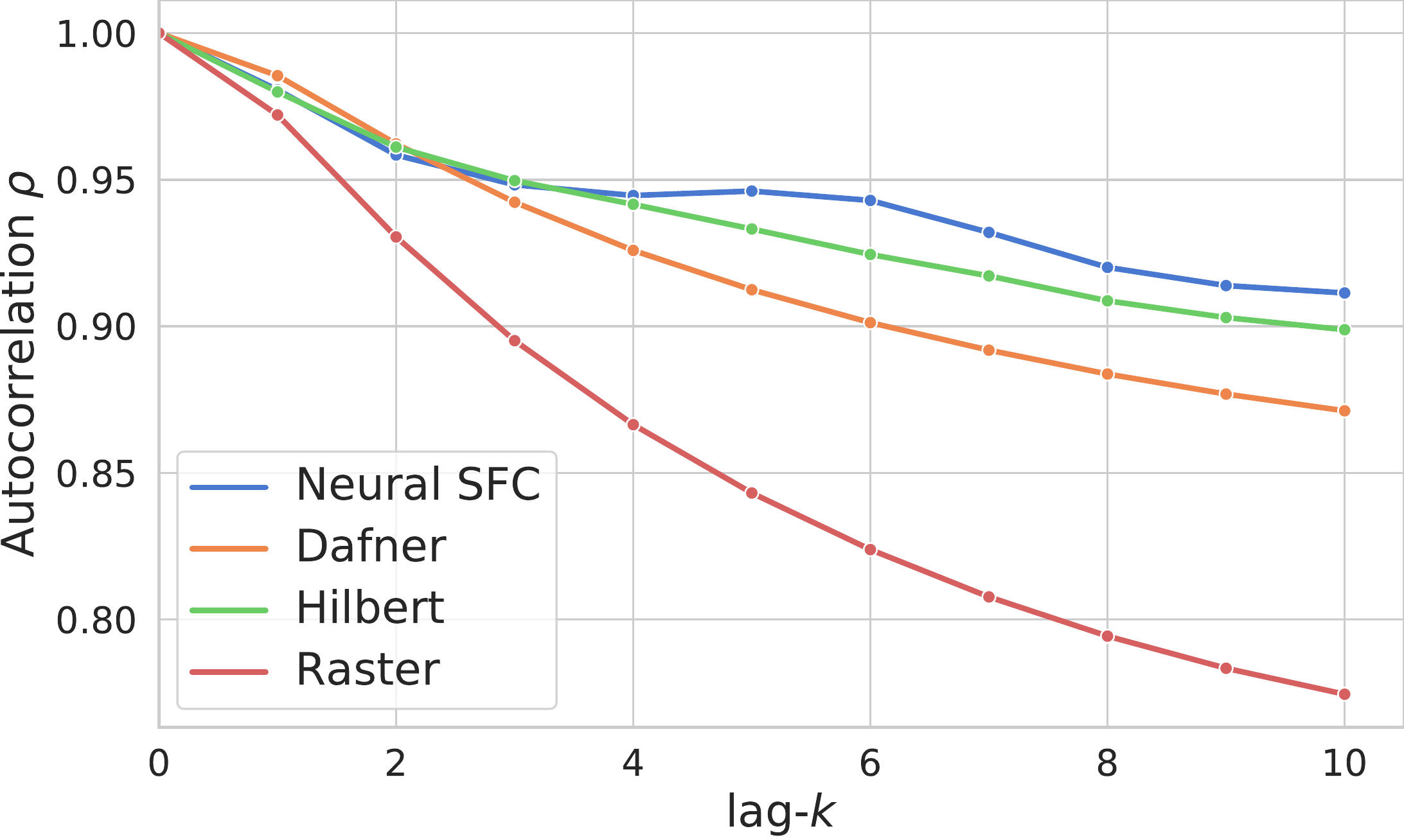}\\
(a) MNIST & (b) Fashion-MNIST & (c) FFHQ $32 \times 32$\\
\end{tabular}
\caption{lag-$k$ autocorrelation for MNIST, Fashion-MNIST and FFHQ datasets. While Dafner \etal provide higher autocorrelation for small lag, \ie, from $k=2$ to $k=4$, Neural SFCs outperform Dafner \etal for $k>4$ in all the datasets. Note that we trained our model for $k=6$, and hence this behaviour is expected.}
\label{fig:ac}
\end{figure*}

\subsection{Optimizing Autocorrelation}
\label{sec:quanteval}

To quantitatively evaluate the generated SFCs, we plot lag-$k$ autocorrelations of pixel sequences obtained from test sets of three different datasets MNIST, Fashion-MNIST, and FFHQ. Note that even though, we trained our models only to optimize the lag-$6$ autocorrelation, we plot them for a range of values of lag-$k$. From \Cref{fig:ac}, we first observe that the trend is somewhat consistent across multiple datasets, even though they are very different in nature. Neural SFC performs the same or slightly worse than other SFCs at lower values of $k$. However, for $k>4$, it outperforms other SFCs by a wider margin. This is intuitive since our model was optimized to increase the autocorrelation for a large value of lag. This also reflects our model's ability to capture long-range and global information. We believe that higher gains can be obtained if we train and test using the same $k$ value.

In the second and the third columns of Table~\ref{table:ac_compression}, we show the autocorrelation values for MNIST, Fashion-MNIST, and FFHQ.
We observe that the performance of Dafner's SFCs is even worse than the Hilbert Curve in most cases. This indicates that the naive average of several good SFCs may not be a good way to compute the SFC for a set of images.

\begin{table}[!t]
\centering
\caption{Comparison of performance of lag-$k$ autocorrelation and LZW Encoding length for different orders. For autocorrelation, we consistently outperform both the universal and context-based SFC computation approaches at high values of $k$. For LZW Encoding length, we measure the average size per frame in bytes as well as the relative improvement compared to the raster scan order, in the case of each of the datasets. 
We consistently outperform compression performance for other order schemes.}
\begin{tabular}{@{}L{\dimexpr.13\linewidth}L{\dimexpr.3\linewidth}C{\dimexpr.1\linewidth}C{\dimexpr.1\linewidth}C{\dimexpr.25\linewidth}@{}}
\toprule
Dataset & Method & $\rho_6$   $\uparrow$ & $\rho_{10}$ $\uparrow$  & Size in bytes ($\Delta$) $\downarrow$ \\
\midrule
\multirow{3}{*}{MNIST} & Raster & 0.206 & 0.102 & 175.4 \\
& Hilbert  & 0.475 & 0.378 & 182.7 (+7.3) \\
& Dafner~\cite{dafner2000context}  & 0.401 & 0.348 & -  \\
& Neural SFC (Ours) & \textbf{0.558} & \textbf{0.451} & \textbf{171.1 (-4.3)} \\
\midrule
\multirow{3}{*}{FMNIST} & Raster & 0.552 & 0.360 & 425.8 \\
& Hilbert & 0.7  23 & 0.647 & 427.3(+1.5) \\
& Dafner~\cite{dafner2000context} & 0.704 & 0.627 & - \\
& Neural SFC (Ours)  & \textbf{0.786} & \textbf{0.705} & \textbf{412.4 (-13.4)} \\
\midrule
\multirow{3}{*}{FFHQ} & Raster & 0.824 & 0.775 & 688.0 \\
& Hilbert  & 0.924 & 0.899 & 689.6(+1.6) \\
& Dafner~\cite{dafner2000context} & 0.901 & 0.871 & - \\
& Neural SFC (Ours)  & \textbf{0.943} & \textbf{0.911} & \textbf{678.3 (-9.7)} \\
\midrule
\multirow{3}{*}{TGIF} & Raster & - & - & 563.9 \\
& Hilbert & - & - & 567.0 (+3.1) \\
& Neural SFC (Ours) & - & - & \textbf{556.9 (-7.0)} \\
\bottomrule
\end{tabular}
\label{table:ac_compression}
\end{table}

\subsection{Optimizing Code Length}
\label{sec:apps}

In this section, we study how Neural SFC can improve image compression results. Specifically, we use the LZW length objective to optimize $\E$. This means we set $\Phi = L$ in Algorith~\ref{alg:training}, where $L$ is defined in Eq.~\ref{eq:lzwl}.
We evaluate our model's performance on all 4 datasets. 
On MNIST and Fashion-MNIST, Neural SFC model is trained for each class label. FFHQ has no labels, therefore all images are used together to learn a Neural SFC. On the TGIF dataset, Neural SFC model is trained on all gifs but generates a unique SFC for each gif, \ie, all frames in a gif share the same Neural SFC.

The last column of Table~\ref{table:ac_compression} shows the average file size required to store an image in the test data and its relative improvement compared to the Raster scan. Specifically, in the TGIF section, the number represents the average size to save a single frame instead of the entire gif file. It is worth noting that this size does not include the order itself. One order can be shared by many images, thus the cost of it can be amortized. We observe that the Neural SFC results in smaller a sequence size on all 4 datasets, showing consistent improvement on the Raster scan or Hilbert Curve. Interestingly, as compared to the Raster scan, even though Hilbert Curve results in higher autocorrelation (see Figure~\ref{fig:ac}), it is always worse in terms of LZW encoding length. This finding suggests that there is no simple relation between an order's autocorrelation performance and LZW encoding length performance.

\subsection{Scaling up SFC}

\begin{figure}[!t]
\centering

\begin{minipage}{.45\linewidth}
\subfloat[Scaling strategy for $2\times2$ grid]{\label{fig:sr1}\includegraphics[width=\textwidth]{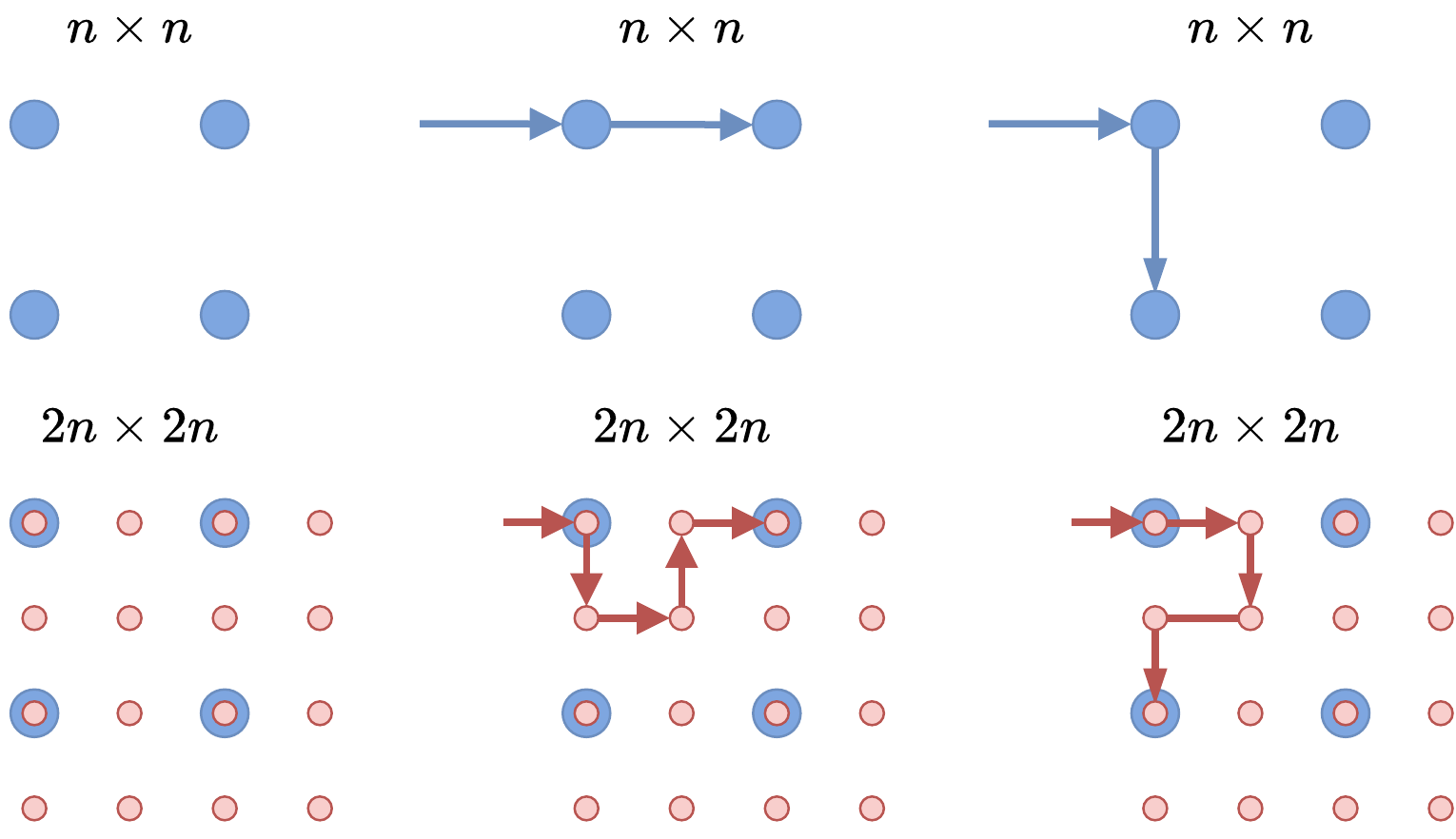}}
\end{minipage} \hfill
\begin{minipage}{.45\linewidth}
\centering
\subfloat[SFC scaled from $5\times5$ resolution to $10\times10$ resolution ]{\label{fig:sr2}\includegraphics[width=\textwidth]{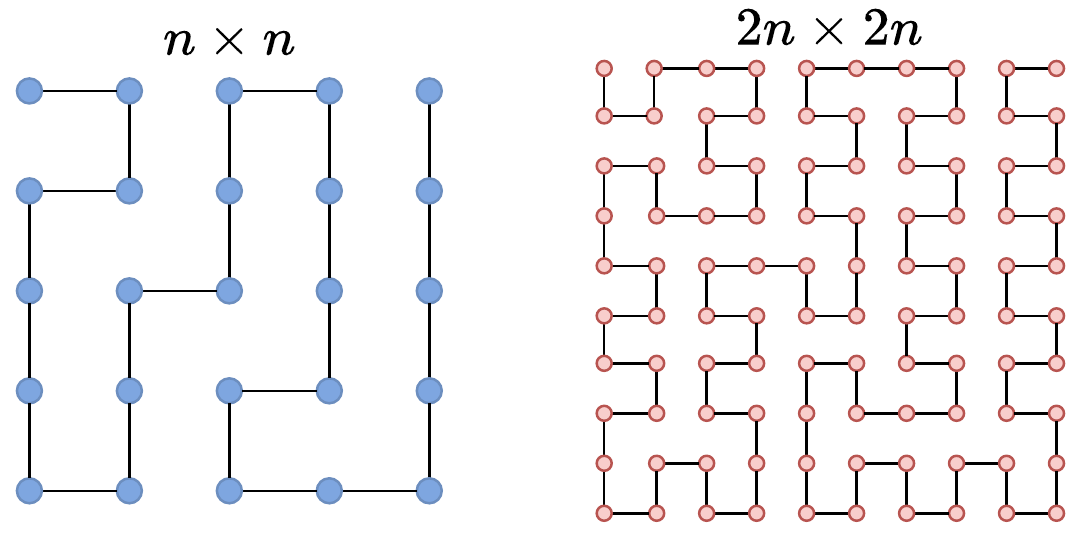}}
\end{minipage}
    \caption{
        \textbf{Scaling up SFC.} The top row (blue circles) shows a $2\times 2$ crop of an image grid of resolution $n \times n$. The bottom row (red circles) shows how we can scale the SFC path from $n \times n$ grid to  $2n \times 2n$ grid. For every incoming SFC to a pixel location, there are 3 possible `next' pixels, straight ahead, left, or right. Column 2 and 3 consider the cases where the SFC goes `straight' or `right'. The image on the right shows a toy example of scaling up SFC for a $5\times5$ image.
    }
    \label{fig:scale}
\end{figure}

Although in the current work, we only train Neural SFC models for $32 \times 32$ images to demonstrate their effectiveness, there are no theoretical restrictions of our approach for higher resolution images. 
It is trivial to either (1) train SFC for higher resolution itself or (2) scale an SFC computed for a low-resolution image to a high-resolution image, preserving the locality properties as shown in \cref{fig:scale}.

\section{Conclusions and Future Work} 
We propose Neural SFC, the first data-driven approach to finding a context-based SFC for a set of images. We parameterize the SFCs as a set of weights over a graph defined using an image grid, and train a neural network to generate the weights. 
Neural SFC can be trained for any objective function defined for a set of images, even when not differentiable. We show the performance of Neural SFC on four real-world datasets on two different objective functions - (1) Pixel autocorrelations in the pixel sequences obtained from an image (2) Compressing a set of images or a short video such as a gif using LZW encoding.

While our work takes the first steps towards finding 1D sequences in 2D data, it opens up a number of directions for future research such as learning SFC for higher dimensional data such as 3D objects or in the case when the 2D space is in a latent space instead of pixels (VQ-VAE~\cite{oord2017neural,razavi2019generating}, dVAE~\cite{ramesh2021zero}). Applying Neural SFCs to large video compression tasks~\cite{ehrlich2022leveraging,chen2021nerv} is also promising given their success on gifs. We will explore these exciting directions and more in future work.

\medskip
\noindent\textbf{Acknowledgements.} This work was partially supported by the Amazon Research Award to AS.

\clearpage
\bibliographystyle{splncs04}
\bibliography{egbib}

\clearpage
\appendix

{\noindent\huge \textbf{Supplementary Material}}

\section{Implementation details}

\paragraph{Neural SFCs.} 
The architectures of the weight generator $\F$, and the weight evaluator $\E$ are shown in Table~\ref{table:architecture_f}, and Table~\ref{table:architecture_e} respectively. Note that the weight evaluator $\E$ takes both an image $I$ and a set of SFC weights $\W$ as inputs. The image $I$ is passed to $\Eenc$ followed by $\EPool$ which computes feature maps $F_\text{map}$. Next, together with the SFC weights $\W$, they are taken as inputs by $\ELine$ to regress the negative autocorrelation.
The number of Residual Blocks and the number of GNN Blocks are denoted by $m_1$ and $m_2$, respectively.
In all our experiments, we use $m_1=8$ and $m_2=6$. GCN~\cite{kipf2016semi} is used as the GNN block for MNIST and Fashion-MNIST datasets. Residual GAT~\cite{vel2018graph} block is used as GNN block for FFHQ and TGIF datasets. More implementation details are available in the code attached.

\begin{table}[h]
    \centering
    \caption{\textbf{Architecture Overview - $\F$}}
    \begin{tabular}{@{}c|c@{}}
        \toprule
        &Weight Generator $\F$  \\
        \midrule
        \multirow{2}{*}{$\Fenc$} & $2 \times 2$ Conv2D (dual graph conv)  \\
        &$m_1 \times$ Residual Block \\
        \midrule
        \multirow{2}{*}{$\FPool$} & Parallel $1 \times 2$ Pooling and $2 \times 1$ Pooling  \\
        & Pooling Results Concatenation \\
        \midrule
        \multirow{1}{*}{$\FLine$} & $m_2 \times$ GNN Block   \\
        \bottomrule
    \end{tabular}
    \label{table:architecture_f}
\end{table}

\begin{table}[h]
    \centering
    \caption{\textbf{Architecture Overview - $\E$}}
    \begin{tabular}{@{}c|c@{}}
        \toprule
        &Weight Generator $\F$  \\
        \midrule
        \multirow{2}{*}{$\Eenc$} & $2 \times 2$ Conv2D (dual graph conv)  \\
        &$m_1 \times$ Residual Block \\
        \midrule
        \multirow{2}{*}{$\EPool$} & Parallel $1 \times 2$ Pooling and $2 \times 1$ Pooling  \\
        & Pooling Results Concatenation \\
        \midrule
        \multirow{4}{*}{$\ELine$} & Addition$($Linear$(\W + F_\text{map}))$  \\
        & $m_2 \times$ GNN Block \\
        & Global Average Pooling \\
        & Linear \& Sigmoid \\
        \bottomrule
    \end{tabular}
    \label{table:architecture_e}
\end{table}

\section{Qualitative Evaluation} 

\begin{figure*}[!ht]
\centering
    \includegraphics[width=.9\linewidth]{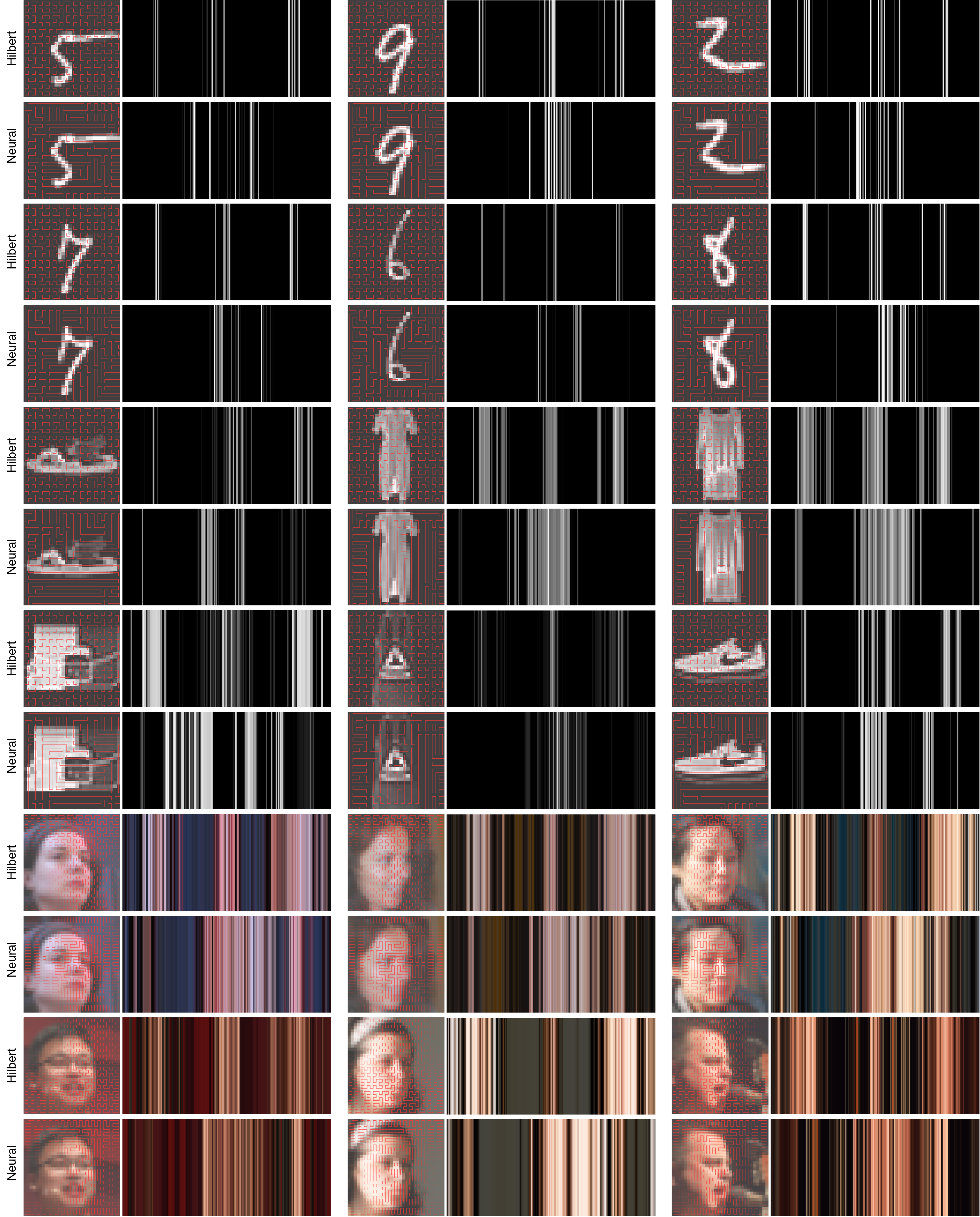}
    \caption{
    Additional qualitative comparison between Hilbert curves and Neural SFCs. Left: SFC (in red color) overlayed on the image. Right: Image flattened according to the SFC and visualized in 1-dimension. Images in the top four rows are from MNIST, the ones in the middle four rows are from Fashion-MNIST, and the ones in the bottom four rows are from FFHQ Faces. Neural SFCs on images from MNIST and Fashion-MNIST are class-conditional, \ie, computed for each class. 
    Best viewed in color.
    }
    \label{fig:vis_more_supp}
\end{figure*}

In \fig~\ref{fig:vis_more_supp}, we show more examples to compare the Neural SFCs with the Hilbert curves. In each image pair, left image shows the original image overlayed with Space-filling Curves the red. The right image shows a 1D representation of the image obtained by just flattening the pixel colors in the SFC order.
Although Neural SFCs are averaged on each class label (MNIST and Fashion-MNIST) or even the entire dataset (FFHQ), it is still clear to see how Neural SFCs keep better long-range spatial coherence than Dafner SFCs.

\section{Quantitative Evaluation}

\begin{figure*}[!h]
    \centering

    \subfloat[MNIST Class 0]{
        \includegraphics[width=.3\textwidth]{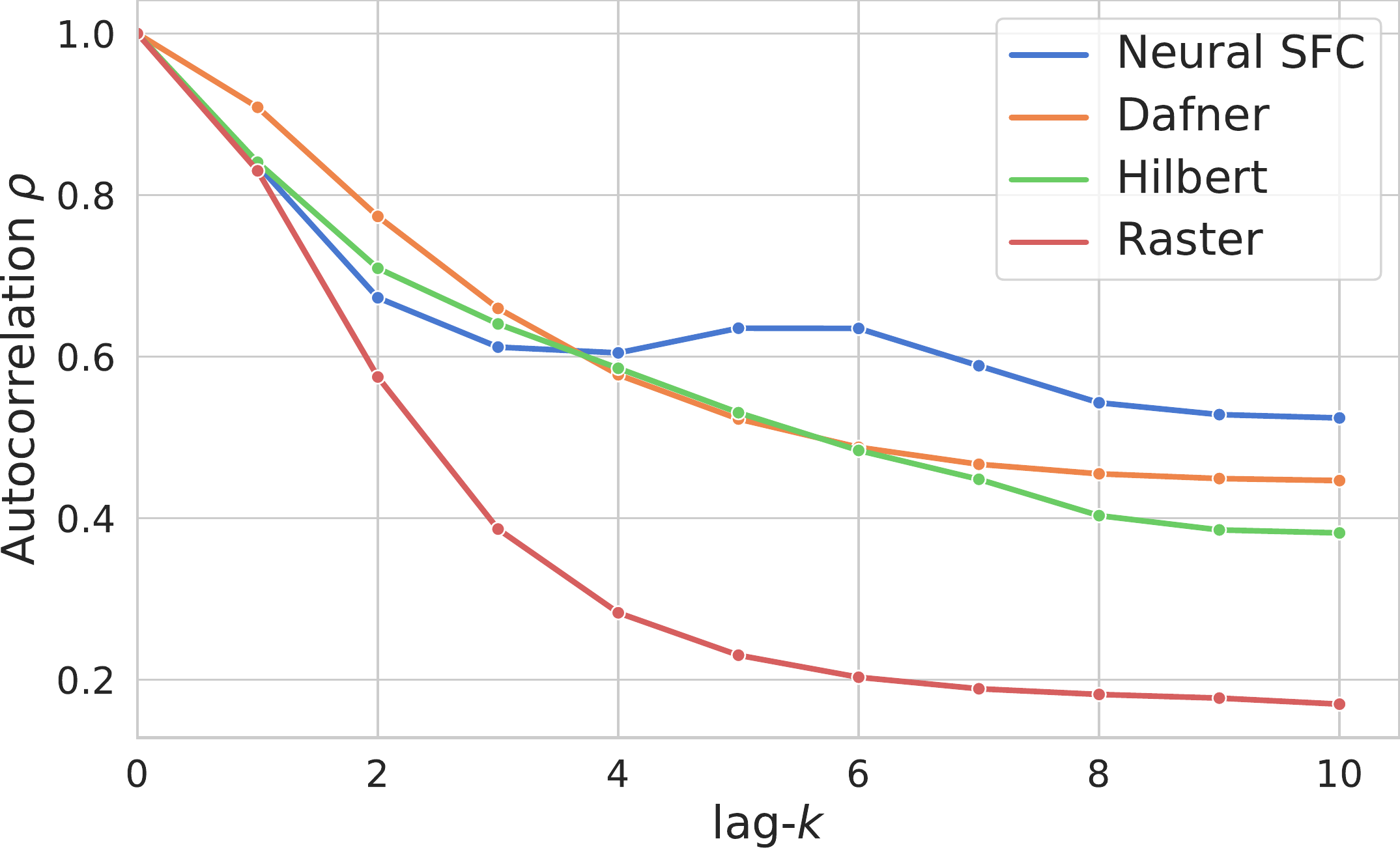}
    }
    \quad\quad
    \subfloat[MNIST Class 1]{
        \includegraphics[width=0.3\textwidth]{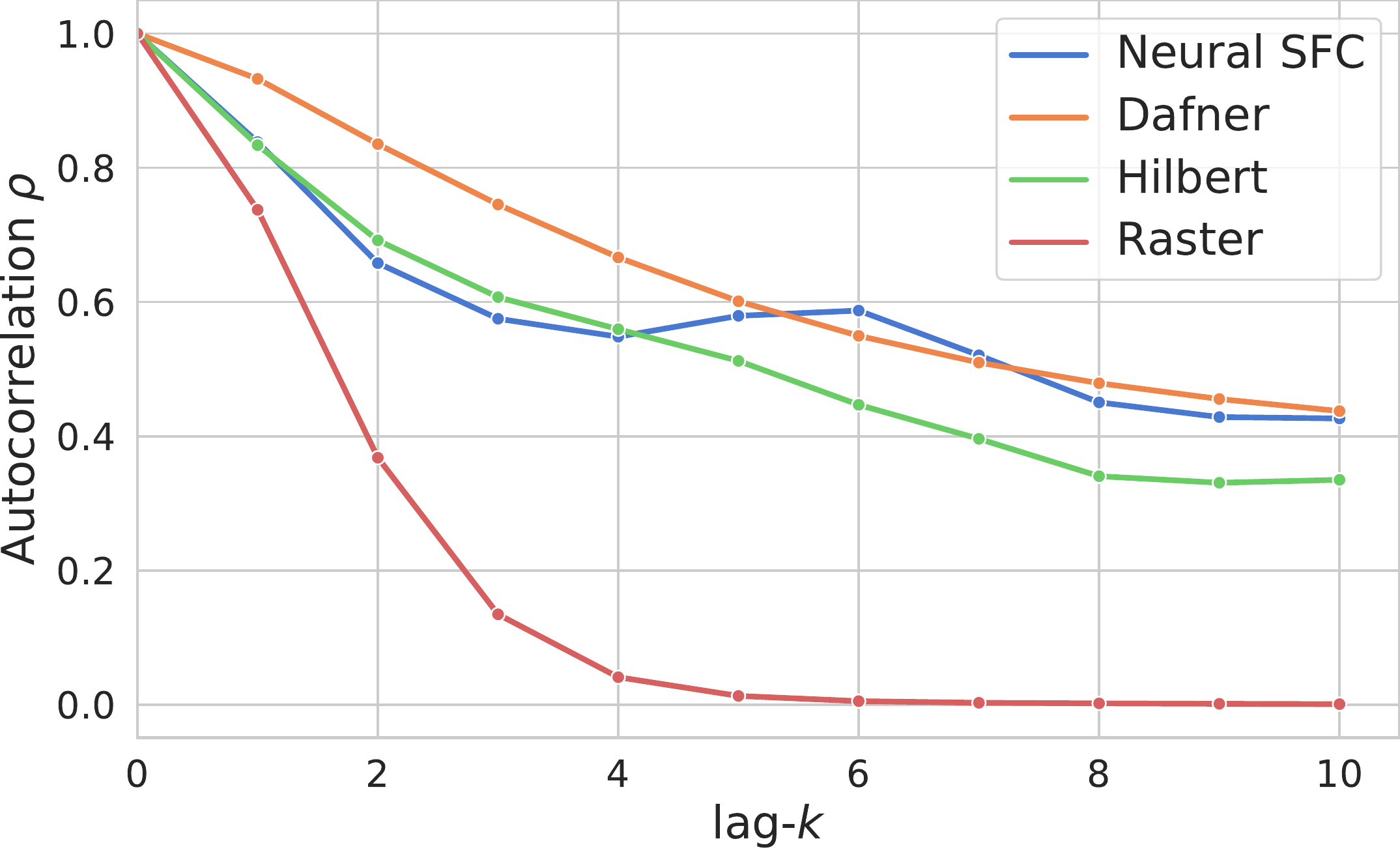}
    }

    \subfloat[MNIST Class 2]{
        \includegraphics[width=.3\textwidth]{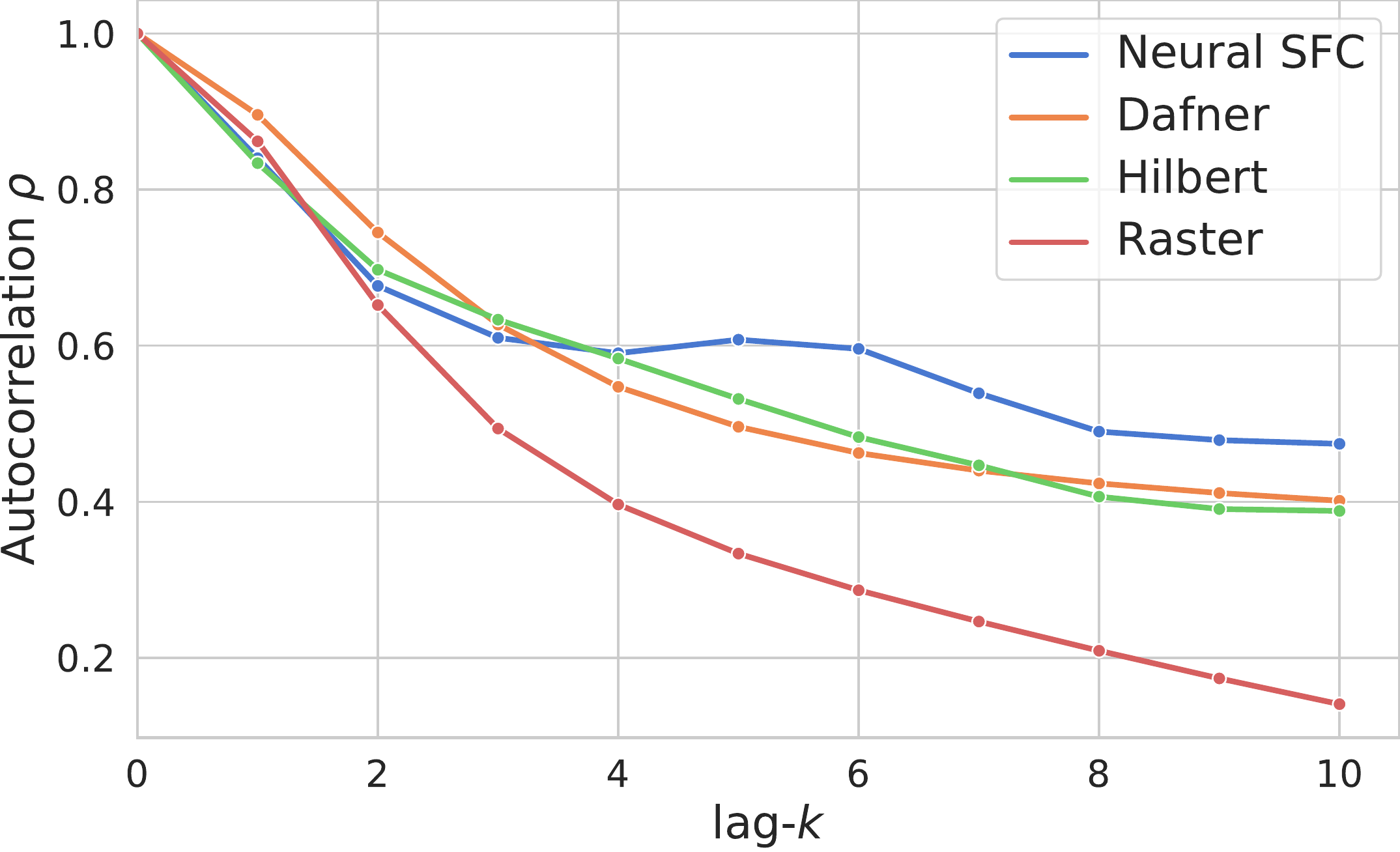}
    }
    \quad\quad
    \subfloat[MNIST Class 3]{
        \includegraphics[width=0.3\textwidth]{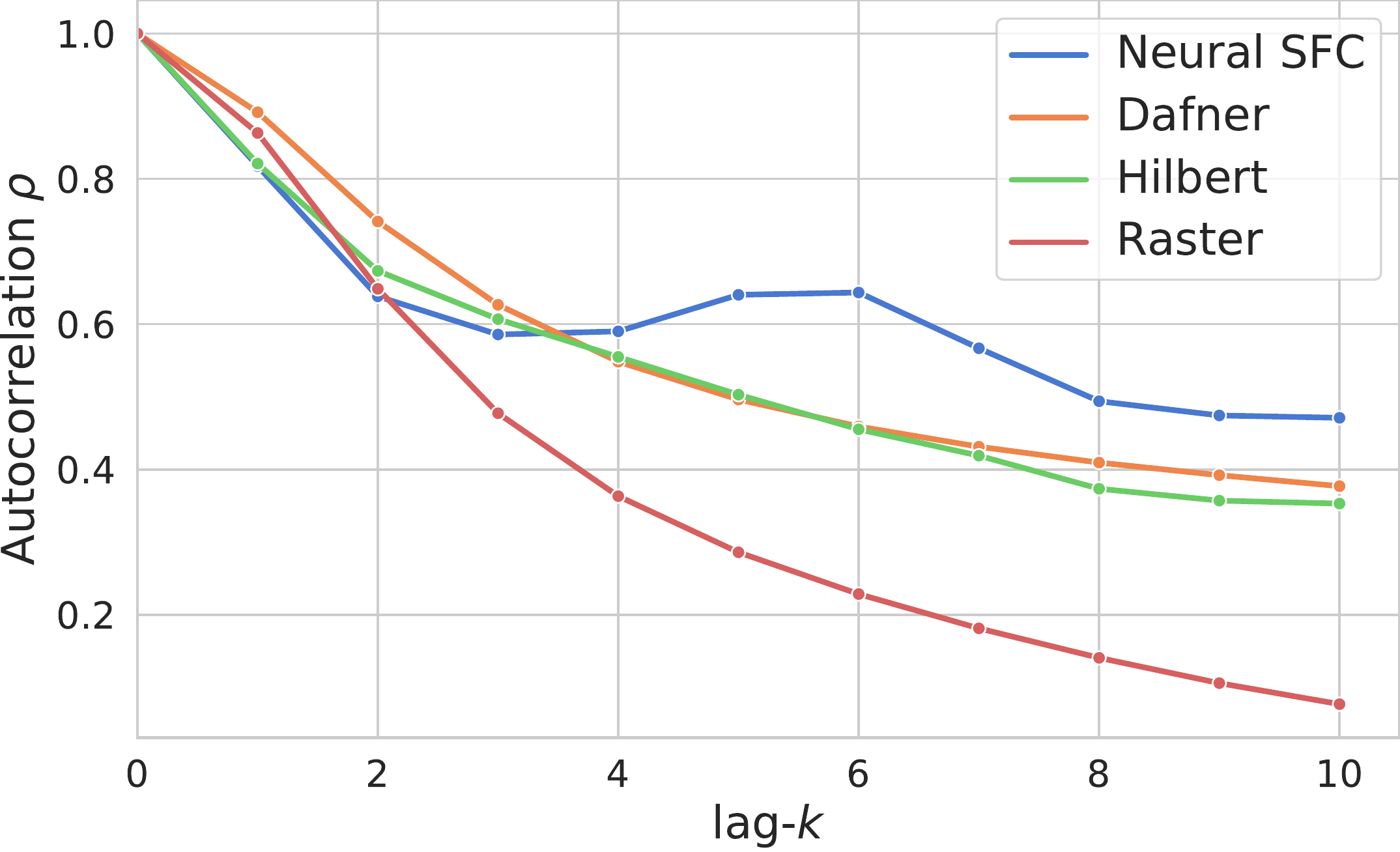}
    }

    \subfloat[MNIST Class 4]{
        \includegraphics[width=.3\textwidth]{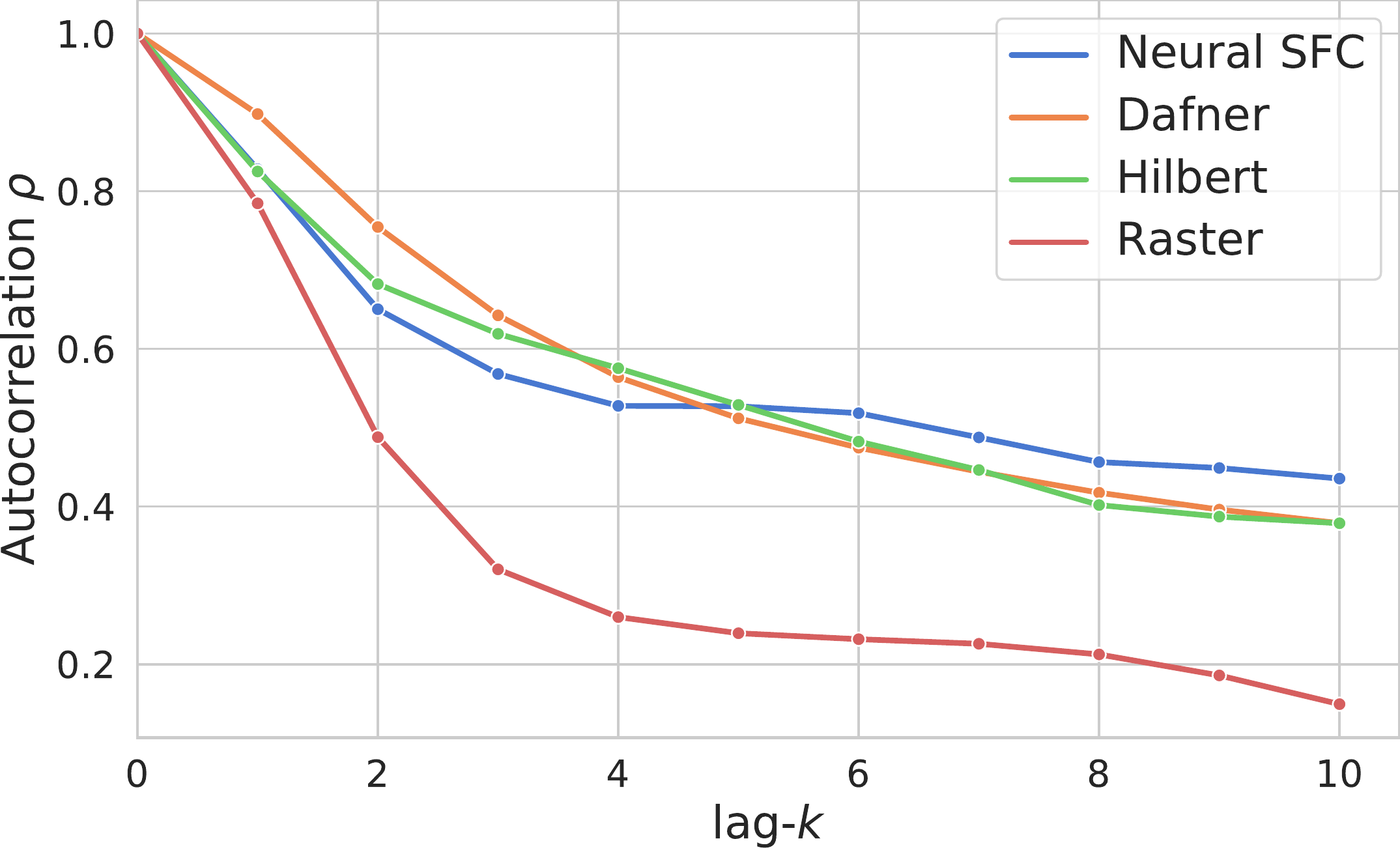}

    }
    \quad\quad
    \subfloat[MNIST Class 5]{
        \includegraphics[width=0.3\textwidth]{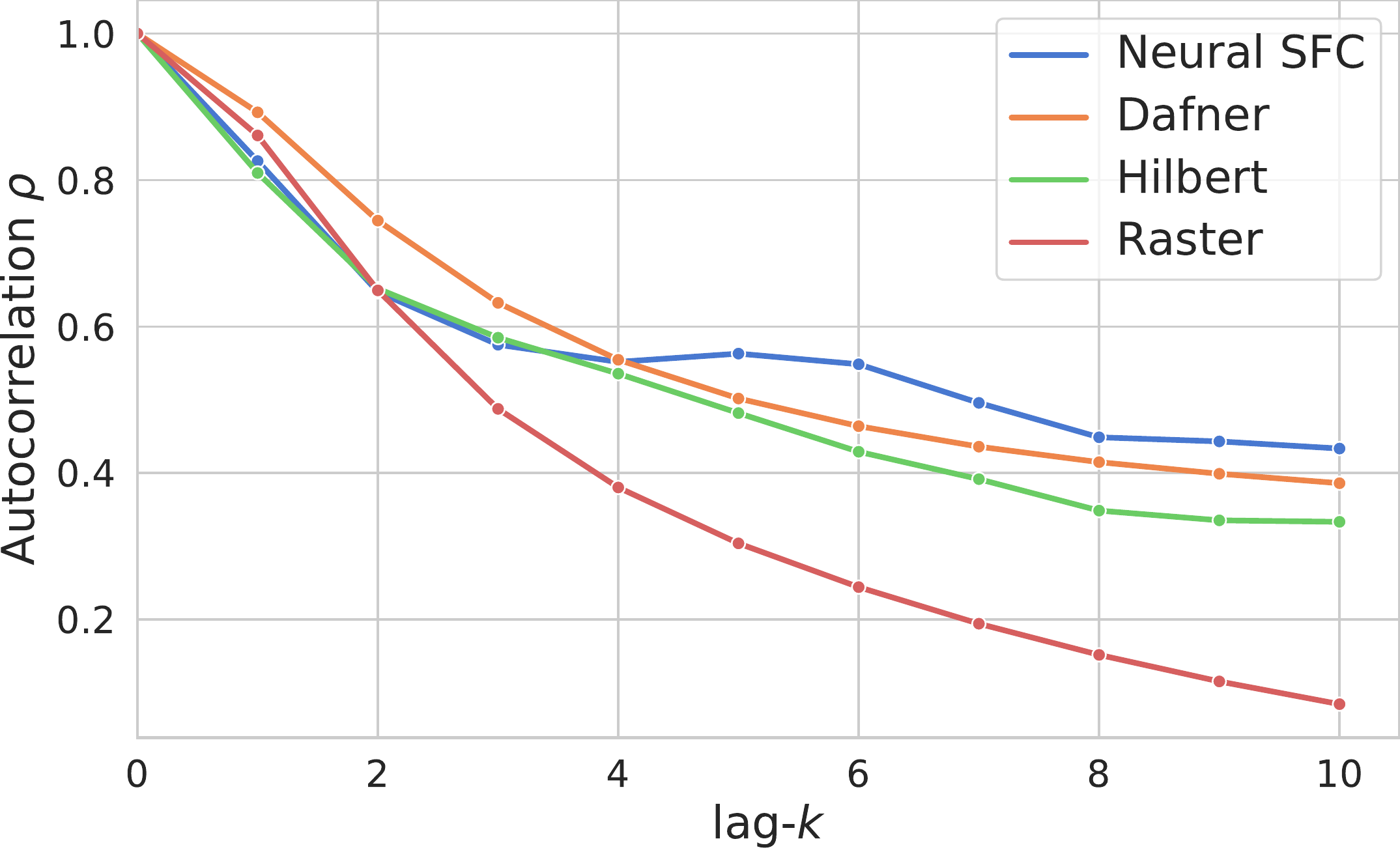}
    }
    
    \subfloat[MNIST Class 6]{
        \includegraphics[width=.3\textwidth]{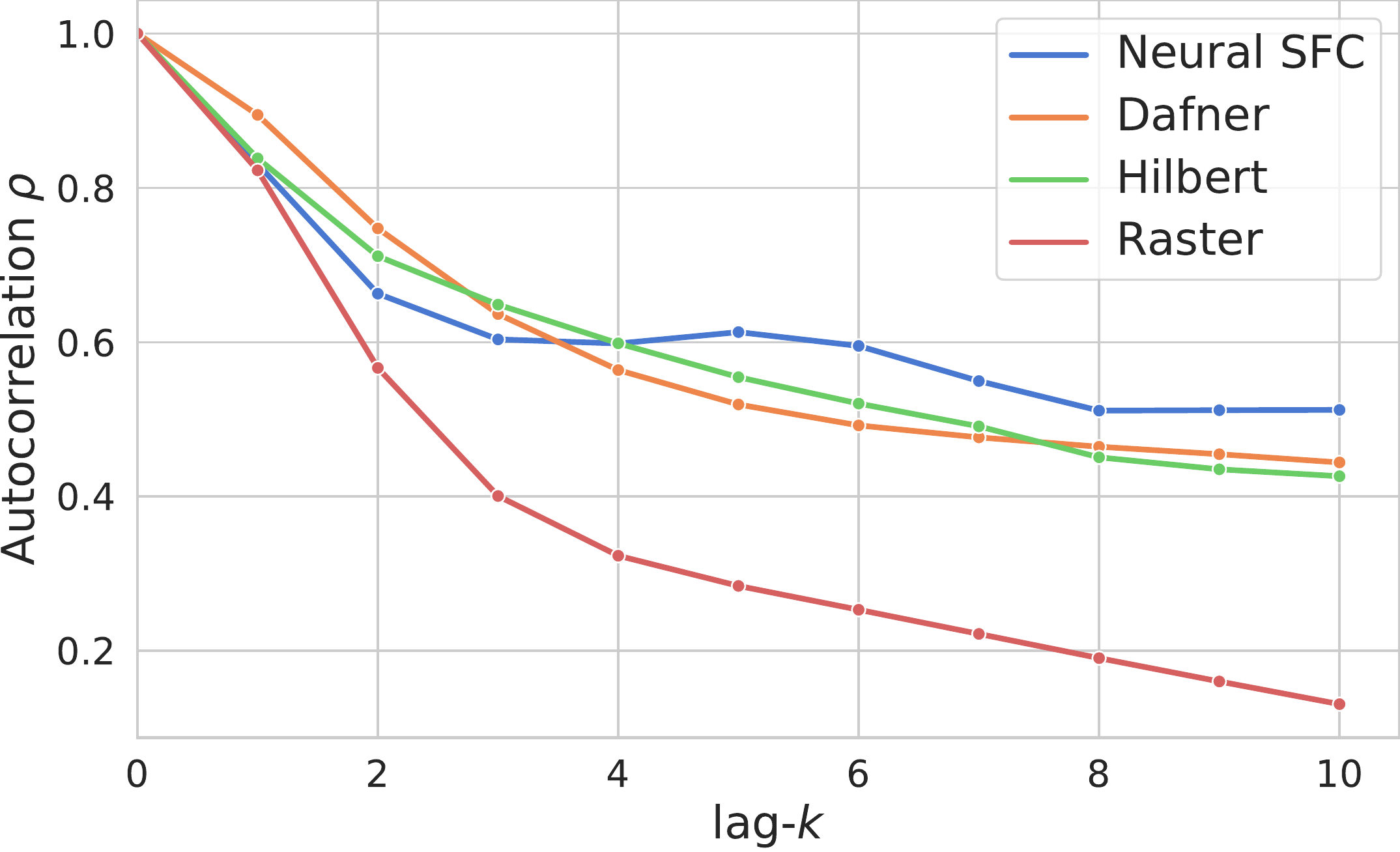}

    }
    \quad\quad
    \subfloat[MNIST Class 7]{
        \includegraphics[width=0.3\textwidth]{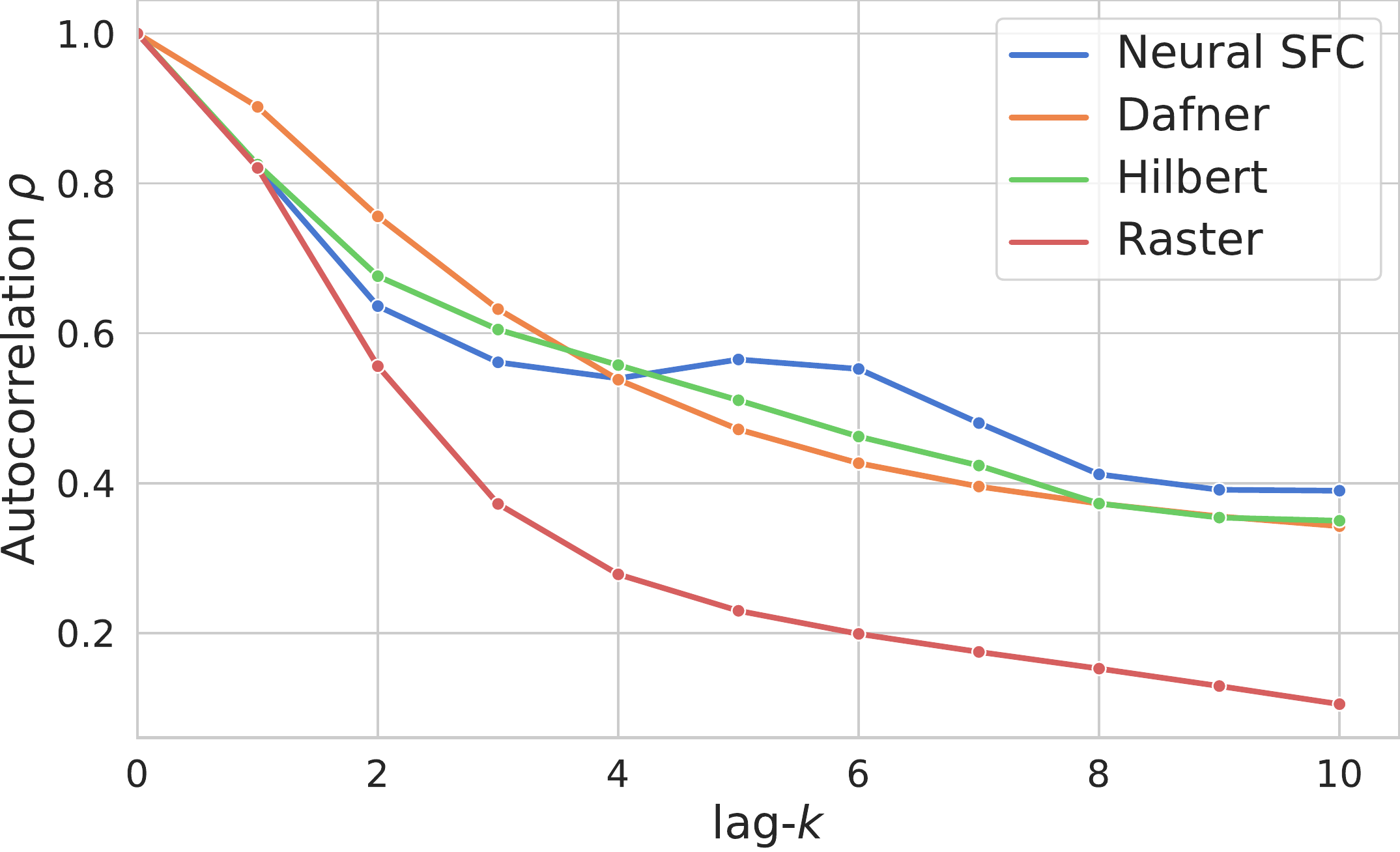}
    }

    \subfloat[MNIST Class 8]{
        \includegraphics[width=.3\textwidth]{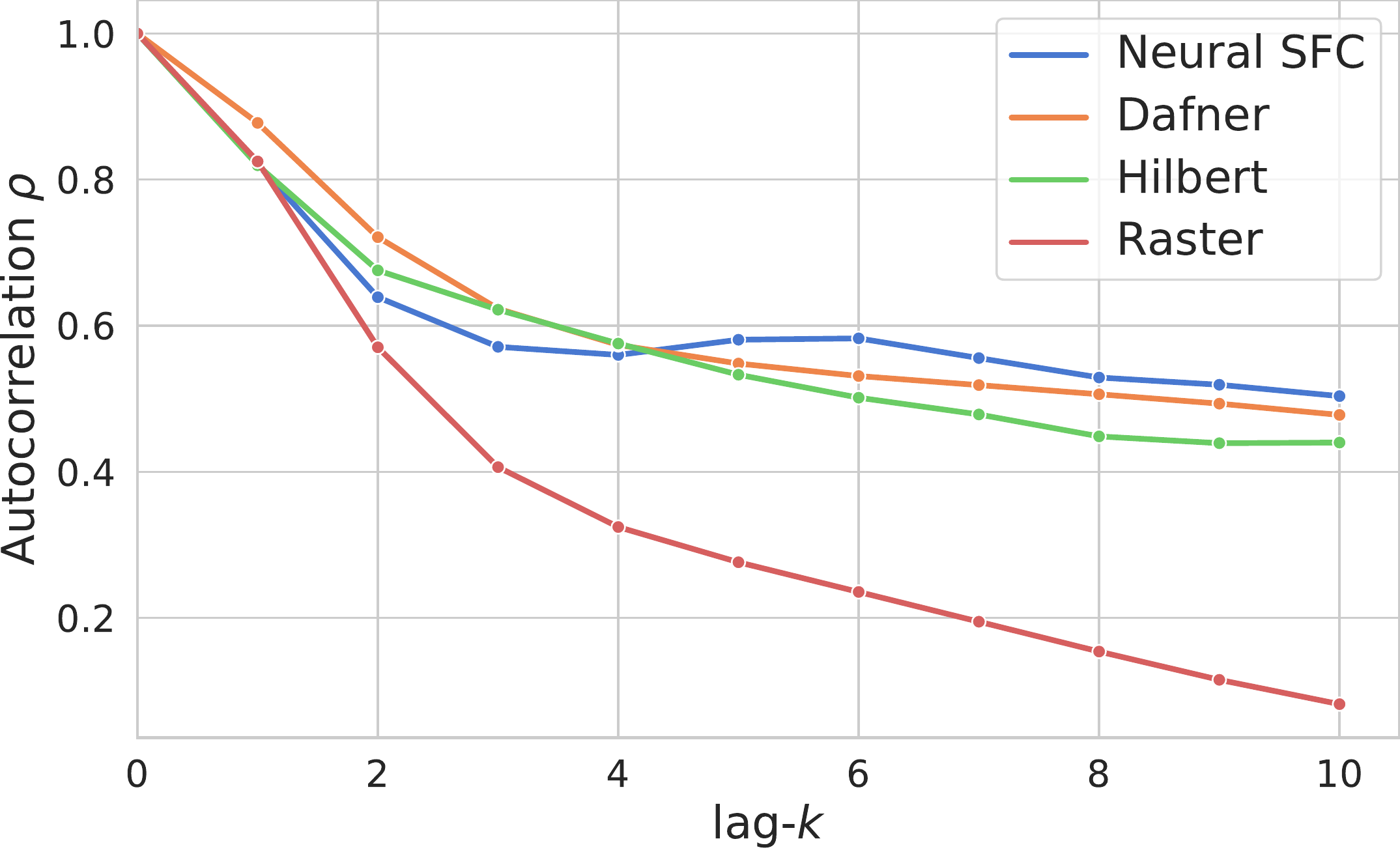}
    }
    \quad\quad
    \subfloat[MNIST Class 9]{
        \includegraphics[width=0.3\textwidth]{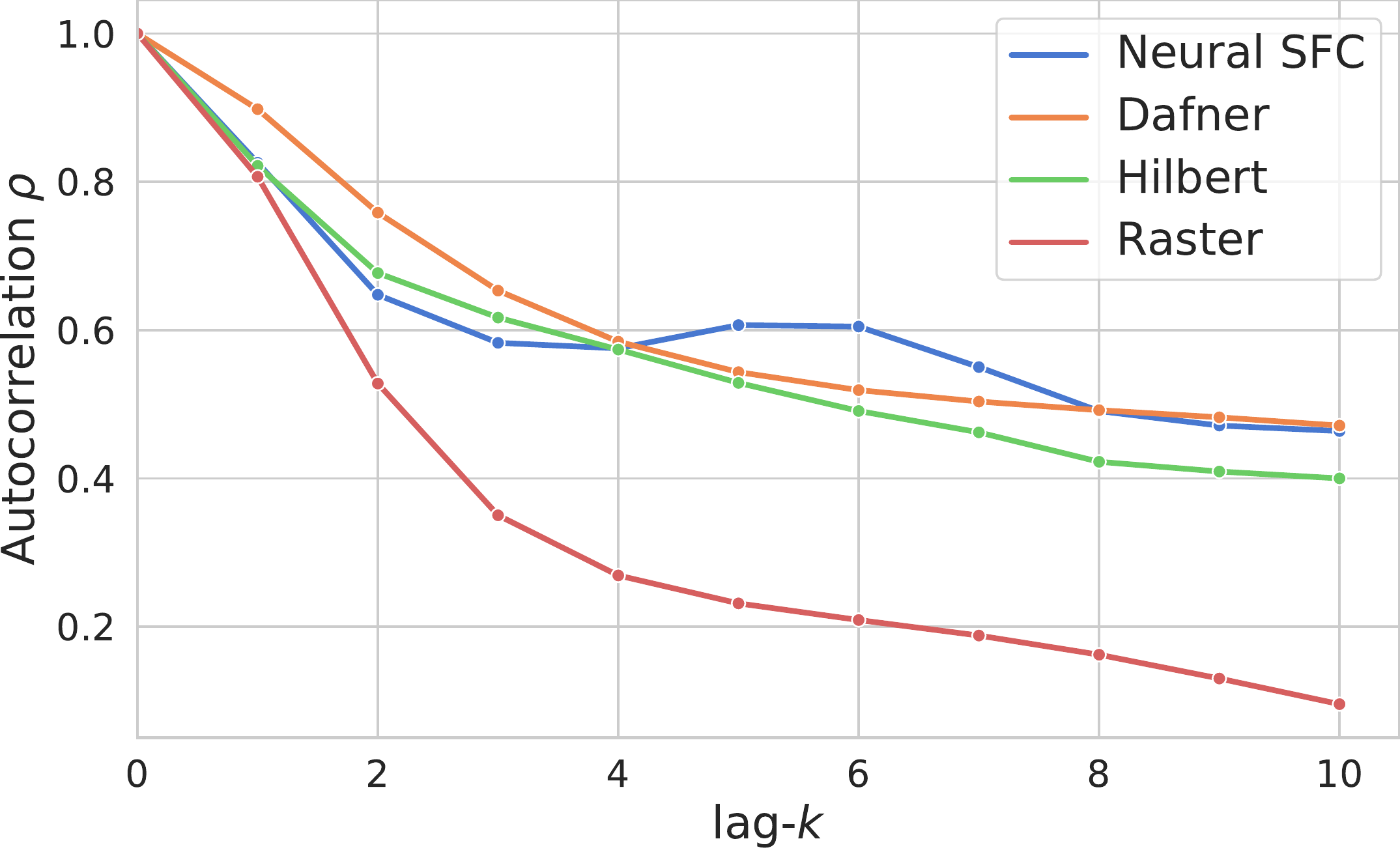}
    }

    \caption{lag-$k$ autocorrelations of SFCs on class conditional MNIST}
    \label{fig:ac_cmnist05}
\end{figure*}

We provide additional autocorrelation results on the class conditional MNIST datasets in~\Cref{fig:ac_cmnist05}. In this case, we train multiple Neural SFC models on the subsets of MNIST corresponding to each class labels separately using the lag-6 autocorrelation objective. Then we evaluate these models on the corresponding test sets. We can observe the similar trends as described in Section~4.4 in the main paper. We observe that Neural SFCs perform the best at $k=6$ which is also the value we used during training. Also that the class conditional image sets are about 10 times smaller than the full MNIST dataset, so it's understandable that the performance of Neural SFCs are not that good on certain subsets. %

\section{Ablation for Lag-$k$ Autocorrelation}
\begin{figure}[!h]
\centering
    \includegraphics[width=0.5\textwidth]{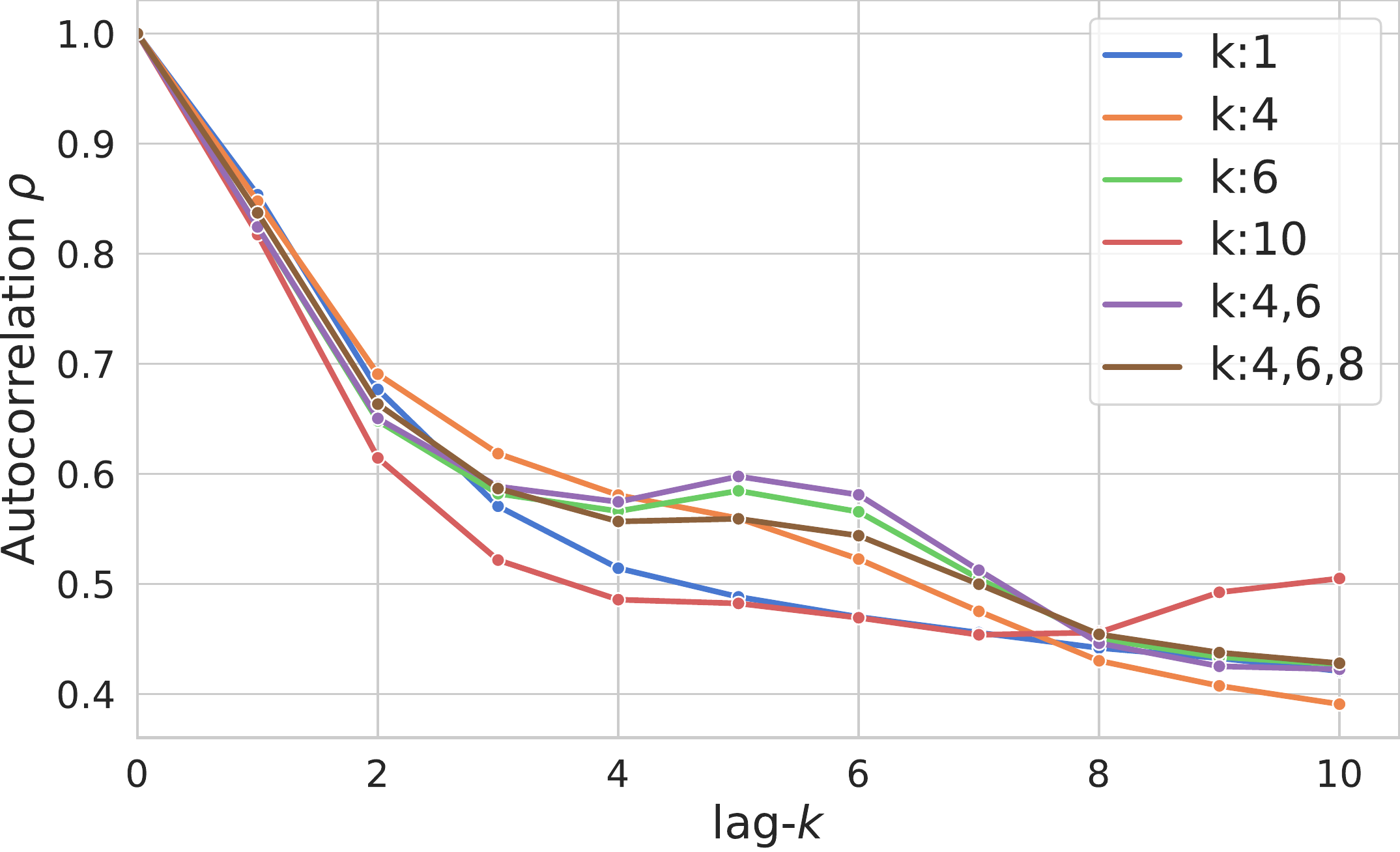}
    \caption{\textbf{Image-set} SFCs with different training $k$ on MNIST}
    \label{fig:k_ablation}
    \vspace{-0.2in}
\end{figure}

In Figure~\ref{fig:k_ablation}, we show lag-$k$ autocorrelation for Neural SFCs trained using different values of $k$ or combinations of different values of $k$. 
When training using multiple values of $k$, the loss values are averaged evenly on them for both the weight generator $\F$ and the weight evaluator $\E$. 
We can see $k=6$ is generally the best choice among all single $k$ training settings.
But if we train NerualSFCs using $k=4, 6$ simultaneously, we can obtain even better autocorrelations from $k=4$ to $6$. %
However, training using $k=4, 6, 8$ results in worse performance.

\section{Performance of the Weight Evaluator} 
\begin{figure}[!ht]
\centering %
\includegraphics[width=0.6\linewidth]{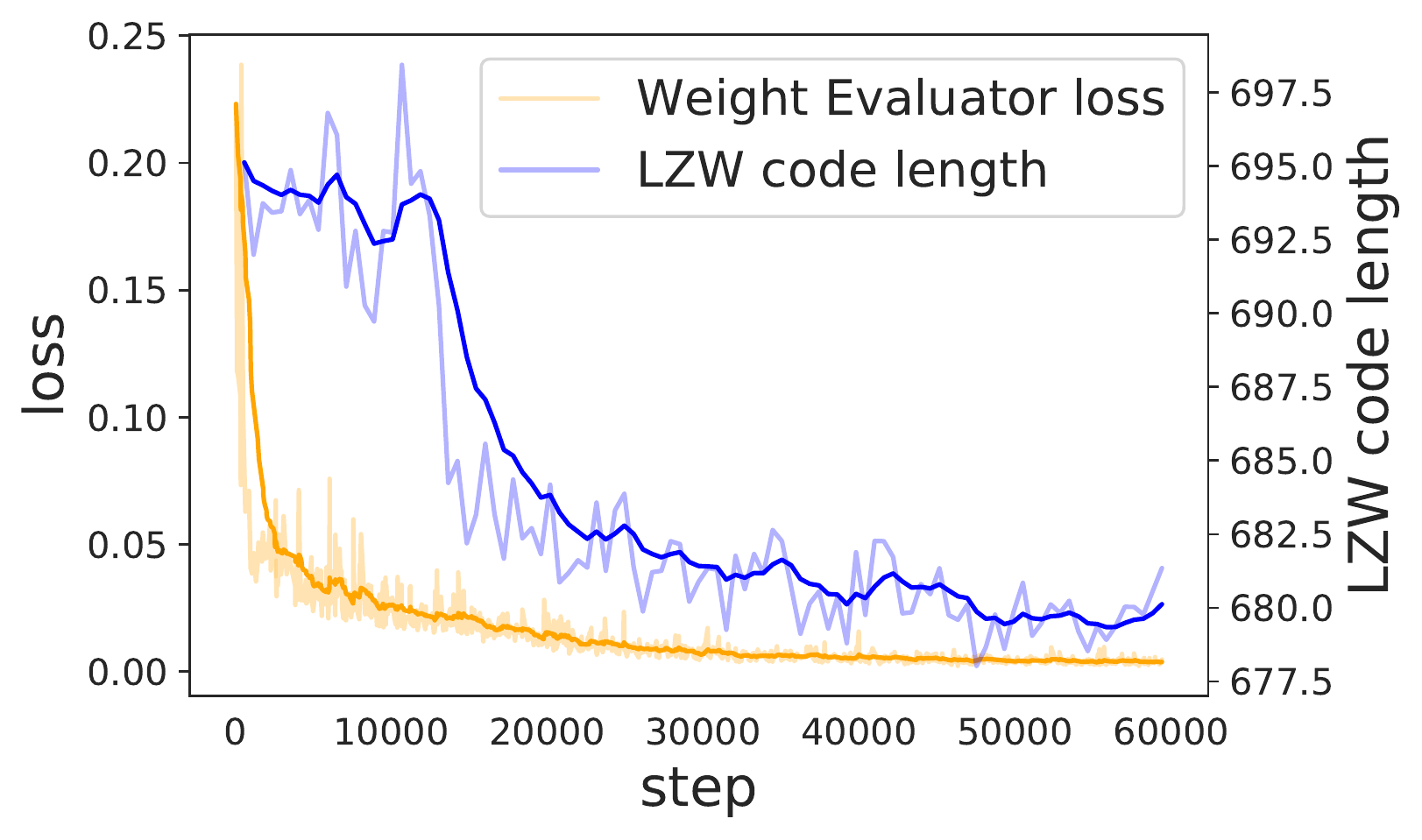}
\vspace{-0.15in}
\caption{Weight Evaluator Loss and the LZW code length.}
\vspace{-0.15in}
\label{fig:e_loss}
\end{figure}

Fig.~\ref{fig:e_loss} shows a typical loss (MSE) of the Weight Evaluator $\E$ and the LZW code length resulting from the Weight Generator during training. In above experiment, NeuralSFC is trained on FFHQ dataset using the LZW code length objective. As the loss values get close to 0, they can provide sufficient signal to guide the weight generator $\F$ which is apparent in Fig.~\ref{fig:e_loss}. 

\end{document}